\pdfoutput=1
\documentclass[11pt]{article}

\usepackage[preprint]{acl}

\usepackage{times}
\usepackage{latexsym}
\usepackage[T1]{fontenc}
\usepackage[utf8]{inputenc}
\usepackage{microtype}
\usepackage{inconsolata}
\usepackage{graphicx}
\usepackage{booktabs}
\usepackage{amsmath}
\usepackage{amssymb}
\usepackage{multirow}
\usepackage{stfloats}
\usepackage{placeins}
\usepackage{algorithm}
\usepackage{algpseudocode}
\usepackage[skins,breakable]{tcolorbox}
\usepackage{enumitem}

\newtcolorbox{promptbox}[1]{%
  enhanced, breakable,
  colback=black!2!white, colframe=black!40!white,
  boxrule=0.45pt, arc=1.5pt,
  left=5pt, right=5pt, top=4pt, bottom=4pt,
  lefttitle=5pt, righttitle=5pt, toptitle=2.5pt, bottomtitle=2.5pt,
  fonttitle=\footnotesize\bfseries, coltitle=black,
  colbacktitle=black!7!white,
  title={#1},
  fontupper=\footnotesize\raggedright
}

\newcommand{\method}{\textsc{Vernier}}
\newcommand{\ul}[1]{\underline{#1}}

\title{Compare Before You Pack: Single-Pass Joint Evidence Selection for Long-Term Conversational Memory}

\author{Beichen Huang, Zongxi Li, Haoran Xie \\
  Division of Artificial Intelligence, Lingnan University, Hong Kong SAR \\
  {\small\textbf{Code:} \url{https://github.com/Beichen-Hg/Vernier} \quad \textbf{Models \& Data:} \url{https://huggingface.co/BertramM}}}

\begin{document}
\maketitle

\begin{abstract}
Long-term conversational agents answer queries over hundreds of sessions where few turns support the answer. Verbatim retrieval scores turns individually for topical relevance, missing state updates, multi-turn links, and cross-query calibration, forcing systems to feed noisy, outdated turns inside a fixed top-$k$ window. Most memory systems therefore rely on offline LLM memory construction to resolve conflicts, link sessions, and compact context, burning tens of millions of tokens and irreversibly losing dates or phrasing. To recover these benefits directly over lossless verbatim retrieval, we propose \textbf{\method{}}, a single-pass evidence selector trained over frozen retrieval. Given a query and retrieved candidates, \method{} compares all turns bidirectionally in one forward pass, retains compact evidence sets, and prunes redundant or outdated turns within a budget under one fixed threshold. Evaluated zero-shot across five readers and four benchmarks up to 1.5M tokens, \method{} beats offline builders by $\ge$+7.7\,pp on \textsc{LongMemEval-S} with zero LLM builds. It also outperforms full-budget reranking by +2.5--2.8\,pp on \textsc{LongMemEval-S/M} using 50\%--57\% fewer tokens, matching end-to-end latency on \textsc{LongMemEval-S}.
\end{abstract}

\begin{figure*}[t]
    \centering
    \includegraphics[width=\textwidth]{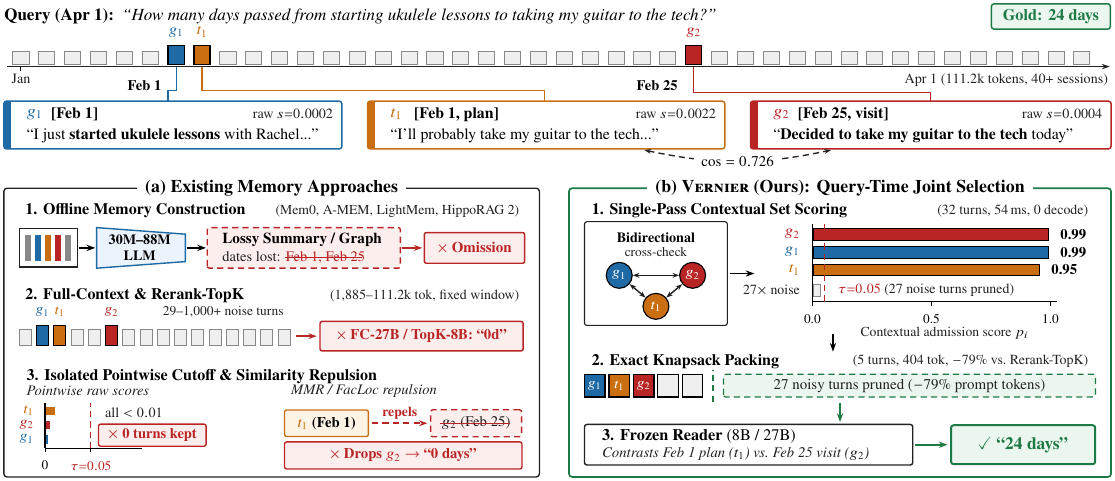}
    \caption{Failure contrast on \textsc{LME-S} \texttt{4dfccbf7}, 111.2k tokens, gold \emph{24 days}. Top: gold turns $g_1$ and $g_2$ sit 24 days apart beside near-duplicate Feb~1 plan $t_1$ ($\cos=0.726$). Panel~(a): (1) offline memory construction spends 30M--88M tokens and loses dates; (2) full context and \textsc{Rerank-TopK} dilute focus across 29--1,000+ noise turns; (3) unfloored pointwise cutoffs drop both golds, while similarity repulsion discards $g_2$. Panel~(b): \method{} scores 32 turns jointly in 54\,ms and packs 5 turns in 404 tokens ($-79\%$) so all readers answer \emph{24 days}.}
    \label{fig:motivated}
\end{figure*}

\section{Introduction}

Long-term conversational assistants must track user preferences, events, and evolving facts across months of dialogue \citep{xu2022goldfish,maharana2024locomo,wu2025longmemeval}. As histories reach hundreds of sessions, reading full dialogue is costly and dilutes attention \citep{liu2024lost,hsieh2024ruler,zhang2024infinitebench,kuratov2024babilong}. Standard verbatim retrieval recalls turns by isolated relevance, leaving superseded states, multi-turn links, and chatter inside a fixed top-$k$ window. To resolve updates, link scattered facts, and compact context, most systems use offline LLM memory construction via summaries, extraction, or graphs \citep{chhikara2025mem0,gutierrez2025hipporag2,fang2026lightmem}. However, offline compression burns tens of millions of tokens and irreversibly drops dates, phrasing, or assistant turns \citep{an2027fidelity,liu2026trenchcoat,zong2026memprice}.

In this work, we ask whether these benefits can be recovered at query time directly over lossless verbatim retrieval. The core difficulty is that deciding what a reader needs is a set-level problem, not a turn-level one. In Figure~\ref{fig:motivated}, counting days between Feb~1 ukulele lessons $g_1$ and a Feb~25 guitar repair $g_2$ recalls both golds beside an earlier Feb~1 \emph{plan} $t_1$ and 29 noise turns. Scored in isolation, $t_1$ ranks over 5$\times$ above $g_2$, so a fixed top-$k$ window misleads the 8B reader into answering ``0 days.'' Thresholding those isolated scores drops both gold turns, while similarity repulsion \citep{carbonell1998mmr,lin2011submodular} mistakes the actual visit $g_2$ for a near-duplicate of $t_1$ and discards it. Answering accurately requires comparing turns against one another to separate a realized visit from a plan, keep complementary anchors together, and prune surrounding chatter.

Existing query-time selectors struggle to make this pool-wide judgment both accurately and efficiently. Fast pointwise pruners, score-gap heuristics, and post-hoc rescalers operate on isolated relevance scores, so they cannot infer whether one turn updates or complements another. Zero-shot LLM filters, listwise rerankers, and iterative memory locators do compare multiple turns \citep{sun2023rankgpt,lee2025setr,packer2023memgpt,wang2026memloc}. However, they either optimize ordinal ranking instead of subset admission or rely on autoregressive decoding and multi-step search that inflate latency. Effective query-time selection therefore requires a lightweight selector that compares the retrieved pool in one forward pass and outputs calibrated scores deciding both which turns to keep and how many to pack.

\begin{figure*}[t]
    \centering
    \includegraphics[width=\textwidth]{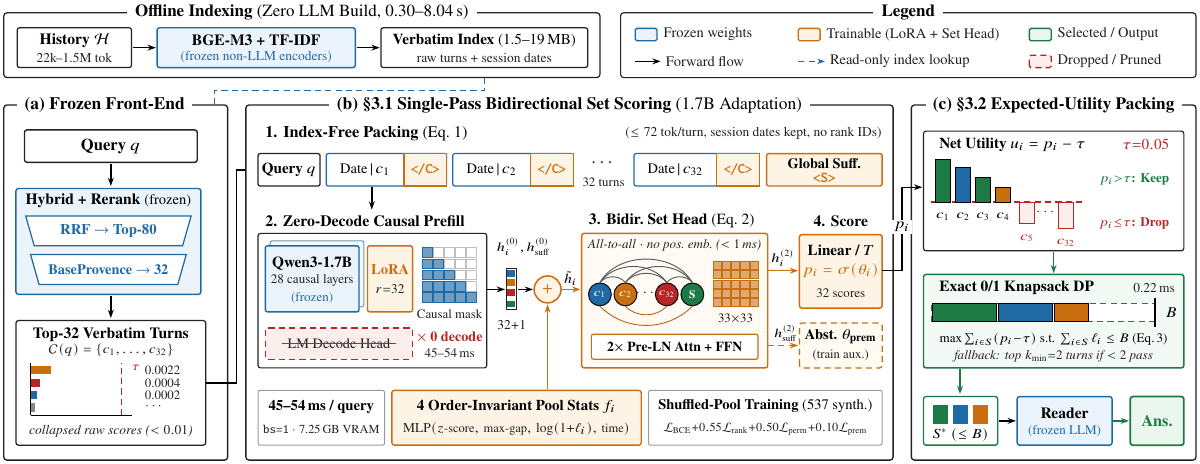}
    \caption{Architecture of \method{}. Top \& (a): frozen \textsc{BGE-M3}+TF-IDF indexing over 22k--1.5M histories and \textsc{BaseProvence} surface Top-32 turns. (b)~\S\ref{sec:set_calib}: four-step \textsc{Qwen3-1.7B} LoRA adaptation runs index-free packing with sentinels, LM-head-free prefill, pool statistics plus a two-layer position-free bidirectional head, and temperature-scaled scoring. (c)~\S\ref{sec:packing}: exact knapsack DP packs $p_i > \tau = 0.05$ within $B$ under a $k_{\min}=2$ fallback.}
    \label{fig:arch}
\end{figure*}

To fill this gap, we introduce \textbf{\method{}}, named after the vernier scale on a caliper that refines a coarse reading into an exact boundary (Figure~\ref{fig:arch}). Operating over coarse frozen retrieval with zero offline LLM builds or online decoding, \method{} selects evidence in two stages. First, a lightweight 1.7B backbone encodes the query and all retrieved turns in one prefill pass. A bidirectional set head then exchanges pool signals symmetrically so early turns are not blind to later distractors (\S\ref{sec:set_calib}). Because retrieval datasets label single-turn relevance rather than cross-turn updates or traps, we train \method{} on QA-grounded synthetic dialogues with minimal gold sets, hard distractors, and empty pools. Second, exact budgeted packing under one shared threshold prunes clutter into the compact 0.6k--0.7k token range of offline builders (\S\ref{sec:packing}).

Our contributions are threefold:
\begin{itemize}
\setlength{\itemsep}{1pt}
\setlength{\parskip}{0pt}
    \item \textbf{Query-time joint evidence selection.} We frame long-term memory selection as calibrated, set-level admission over lossless verbatim retrieval, bridging zero-build fidelity with the conflict resolution and compactness of offline memory construction.
    \item \textbf{\method{} selector and pattern-driven supervision.} \method{} unifies single-pass joint encoding, bidirectional pool exchange, QA-grounded conflict and linking supervision, and exact packing without offline LLM builds or decoding.
    \item \textbf{Empirical frontier and boundaries.} Across five readers and four benchmarks up to 1.5M tokens, \method{} beats offline builders and full-budget reranking at half the context while charting order stability and dense limits.
\end{itemize}

\section{Related Work}

\paragraph{Verbatim vs.\ Structured Long-Term Memory.}
Offline systems invoke LLMs to summarize dialogues, segment episodes, or build graphs \citep{park2023generative,zhong2024memorybank,lee2024readagent,chhikara2025mem0,rasmussen2025zep,xu2025amem,pan2025secom,kang2025memoryos,gutierrez2025hipporag2,fang2026lightmem,liu2026simplemem}, while parametric models compress history into weights \citep{behrouz2025titans}. However, irreversible compression before queries arrive can discard needed details \citep{zong2026memprice}, and verbatim storage often outperforms lossy extraction \citep{an2027fidelity,liu2026trenchcoat}. Even hybrid systems like \textsc{MemLoc} \citep{wang2026memloc}, which feed raw sessions to the reader instead of summaries, still use an offline LLM index and multi-step generation to locate sessions; \method{} indexes without LLMs and selects verbatim turns in one pass.

\paragraph{Adaptive Context Selection and Budgeted Packing.}
Query-time systems retrieve via sparse, dense, or late-interaction search \citep{robertson2009bm25,karpukhin2020dpr,khattab2020colbert,lewis2020rag} or iterative search \citep{packer2023memgpt,asai2024selfrag,qian2025memorag}, then trim text to fit a compact context. Pointwise compressors \citep{jiang2023llmlingua,jiang2024longllmlingua,xu2024recomp,chirkova2025provence} and score-gap cutoffs \citep{taguchi2025adaptivek,derehag2026smartsearch} prune turns without pool-wide comparison. To select subsets jointly, MMR \citep{carbonell1998mmr}, \textsc{FacLoc} \citep{lin2011submodular}, and DPP \citep{kulesza2012dpp} penalize similarity and drop complementary turns, whereas \textsc{SetR} \citep{lee2025setr} relies on autoregressive LLM decoding across both its selection-only and chain-of-thought variants. Other budgeted frameworks act at write time \citep{zong2026memprice} or explore offline hierarchies near 4k tokens \citep{sun2026bear}; \method{} instead acts at query time, packing compact 0.6k--0.7k contexts via exact budgeted dynamic programming \citep{kellerer2004knapsack}.

\paragraph{Contextual Ranking and Label-Supervised Prefill.}
Listwise and setwise rerankers score pools jointly \citep{pang2020setrank,sun2023rankgpt,pradeep2023rankzephyr,reddy2024first,schlatt2025setencoder,huang2026tssetrank,li2026qrranker}, while recent work mitigates position bias via logit debiasing \citep{lv2026capcal} or candidate-isolated shared RoPE \citep{bito2026invarirank}. However, they optimize ordinal ranking, require decoding, or isolate candidates. To eliminate decoding altogether, \textsc{LS-BiLLMs} \citep{li2026lsbillms} discard the LM head, remove causal masks for bidirectional encoding, and train LoRA adapters in label space. \method{} extends label-supervised prefill to multi-candidate scoring with an intact causal backbone and a bidirectional head inspired by set networks \citep{zaheer2017deepsets,lee2019settransformer}. Rather than stopping at ordinal ranks, it pairs these contextual scores with calibration and selective prediction \citep{platt1999probabilistic,zadrozny2002isotonic,guo2017calibration,brier1950verification,chow1970reject,elyaniv2010selective,angelopoulos2024conformal}.

\section{Method}
\label{sec:method}

Given a query $q$ and verbatim history $\mathcal{H} = \{t_1, \dots, t_N\}$ with turn token lengths $\ell_j$, we select a compact subset $S^* \subseteq \mathcal{H}$ within budget $B$ for a frozen reader. Offline indexing uses zero LLMs, encoding verbatim turns and session dates with frozen \textsc{BGE-M3} \citep{chen2024bge} and TF-IDF. At query time, Reciprocal Rank Fusion \citep{cormack2009rrf} retrieves 80 turns and frozen \textsc{Provence} (\textsc{BaseProvence}, \citealp{chirkova2025provence}) trims them to $M = 32$ candidates $\mathcal{C}(q) = \{c_1, \dots, c_M\}$. Because raw pointwise probabilities often collapse below usable cutoffs (Table~\ref{tab:calib_full}), stopping here forces standard pipelines such as \textsc{Rerank-TopK} to fill the entire budget $B$. \method{} operates on this fixed Top-32 pool in two stages: single-pass bidirectional set scoring (\S\ref{sec:set_calib}) and budget-constrained expected-utility packing (\S\ref{sec:packing}).

\subsection{Single-Pass Bidirectional Set Scoring}
\label{sec:set_calib}

To compare all $M=32$ candidates jointly at prefill speed, we adapt \textsc{Qwen3-1.7B} \citep{yang2025qwen3} via rank-32 LoRA \citep{hu2022lora} in four steps (Figure~\ref{fig:arch}, center). \textbf{Step 1: Index-free packing.} We concatenate $q$ and all $M$ turns, capping each at 72 tokens with session dates preserved and numeric rank identifiers omitted to reduce reliance on upstream retrieval rank cues. Each turn ends with sentinel $\texttt{</C>}$, followed by a global sufficiency sentinel $\texttt{<S>}$:
\begin{equation}
    X(q,\mathcal{C}) = [q,\, c_1\texttt{</C>},\, \dots,\, c_M\texttt{</C>},\, \texttt{<S>}].
    \label{eq:pack_seq}
\end{equation}
\textbf{Step 2: Intact zero-decode causal prefill.} Following \textsc{LS-BiLLMs} \citep{li2026lsbillms}, we discard the LM head and run one forward prefill over Eq.~\eqref{eq:pack_seq} to read out candidate states $h_i^{(0)} \in \mathbb{R}^d$ at $\texttt{</C>}$ and global state $h_{\text{suff}}^{(0)}$ at $\texttt{<S>}$. Unlike single-sequence extraction, one pass encodes cross-turn context across all 32 candidates and $\texttt{<S>}$ jointly. Unlike \textsc{LS-BiLLMs}, we keep pretrained causal attention intact across all 28 layers to avoid disrupting the backbone under lightweight LoRA and preserve 45--54\,ms prefill speed. Bidirectional exchange instead operates at the 33-node candidate level. \textbf{Step 3: Pool-statistics fusion and bidirectional exchange.} To counter causal front-position asymmetry, we compute four order-invariant statistics $f_i$, encoding reranker $z$-score, gap to pool max, $\log(1+\ell_i)$, and normalized session time, and fuse $\tilde{h}_i = h_i^{(0)} + \mathrm{MLP}(f_i)$ via a residual MLP. All 32 candidates and $h_{\text{suff}}^{(0)}$ then enter a two-layer Pre-LN bidirectional Transformer \emph{without candidate positional embeddings} \citep{lee2019settransformer}:
\begin{equation}
    \bigl[h_{\text{suff}}^{(2)}, \{h_i^{(2)}\}\bigr] = \mathrm{Bidir}_2\bigl([h_{\text{suff}}^{(0)}, \{\tilde{h}_i\}]\bigr),
    \label{eq:bidir}
\end{equation}
exchanging dates, updates, and competition symmetrically across 33 nodes in under 1\,ms. \textbf{Step 4: Temperature-scaled admission scoring.} A shared linear head with learned temperature $T$ maps $h_i^{(2)}$ to logit $\theta_i = (w^\top h_i^{(2)} + b)/T$ and score $p_i = \sigma(\theta_i)$, while a parallel head on $h_{\text{suff}}^{(2)}$ outputs unanswerable-query logit $\theta_{\text{prem}}$.

\paragraph{Necessity of Training and Pattern-Driven Supervision.}
Training is necessary because post-hoc rescalers cannot infer cross-turn relations from isolated scores, and zero-shot LLM gating decodes slowly without reliable cutoffs (\S\ref{sec:rq2}). Because we train \method{} for the update-disambiguation, multi-session linking, and clutter-pruning patterns of Figure~\ref{fig:motivated}, standard retrieval datasets fall short: they label query relevance rather than minimal answer-supporting turns versus traps. To avoid mimicking target benchmarks, we synthesize 537 QA-grounded dialogues with \textsc{DeepSeek-V4.1-Flash} across six everyday domains to supervise answer sufficiency, trap discrimination, and abstention (Appendix~\ref{app:impl}). We route all 4,622 training queries from 459 dialogues, plus 78 held-out validation dialogues, through the identical frozen Top-32 front-end, mixing short and long turns to block length shortcuts. Using the deployed front-end exposes real distractors and score collapse while reducing any history length to a domain-agnostic 32-turn comparison task. Matched controls trained on the identical synthetic pools still lag without bidirectional set comparison (\S\ref{sec:rq2}), supporting the contribution of candidate interaction rather than synthetic style alone. Only LoRA and the set head train over gold labels $y_i$, hard traps $y_i^{\text{hard}}$, and 10\% empty-pool flags $y_{\text{abs}}$, shuffled per batch under:
\[
    \mathcal{L} = \mathcal{L}_{\text{BCE}} + 0.55\,\mathcal{L}_{\text{rank}} + 0.50\,\mathcal{L}_{\text{perm}} + 0.10\,\mathcal{L}_{\text{prem}}.
\]
Weighted label-smoothed $\mathcal{L}_{\text{BCE}}$ upweights golds and hard distractors to shape threshold-ready admission scores, while softplus pairwise margin $\mathcal{L}_{\text{rank}}$ widens gold-versus-trap separation. Because the lower backbone remains causal, $\mathcal{L}_{\text{perm}}$ runs a second random shuffle every third micro-batch to penalize score drift. Finally, $\mathcal{L}_{\text{prem}}$ supervises $\theta_{\text{prem}}$ against $y_{\text{abs}}$ for abstention (formulas in Appendix~\ref{app:impl}).

\subsection{Budget-Constrained Expected-Utility Packing}
\label{sec:packing}

Treating admission scores $\{p_i\}_{i=1}^M$ as evidence probability proxies motivates an expected-utility rule over lengths $\{\ell_i\}_{i=1}^M$. Let an answer-supporting turn yield reader reward $R > 0$ and a distracting turn incur cost $D > 0$. Approximating set utility additively across turns gives expected net utility $\mathbb{E}[u_i] = p_i R - (1-p_i)D = (R+D)(p_i - \tau)$ with threshold $\tau = D/(R+D) \in [0, 1)$ \citep{chow1970reject}. Maximizing total net utility over binary keep-or-drop decisions within budget $B$ yields an exact knapsack problem \citep{kellerer2004knapsack}:
\begin{equation}
\begin{aligned}
    S^* &= \arg\max_{S \subseteq [M]} \textstyle\sum_{i \in S} (p_i - \tau) \\[-2pt]
        &\quad \text{s.t.}\; \textstyle\sum_{i \in S} \ell_i \le B.
\end{aligned}
\label{eq:knapsack}
\end{equation}
A 0.22\,ms CPU dynamic program solves Eq.~\eqref{eq:knapsack}, and \S\ref{sec:rq2} shows contextualized scores suffice for this additive surrogate. Turns with $p_i \le \tau$ are pruned automatically, shrinking sparse or unanswerable prompts. If fewer than $k_{\min}=2$ turns satisfy $p_i > \tau$, a fallback bypasses DP and greedily selects up to $k_{\min}$ highest-scoring turns within $B$ so readers still inspect local context before abstaining. We lock $\tau = 0.05$ once on synthetic validation, transfer it across all benchmarks, and render turns chronologically in each benchmark's standard format.

\begin{table*}[t]
\centering
\setlength{\tabcolsep}{4pt}
\resizebox{\textwidth}{!}{%
\begin{tabular}{@{}lc|cccc|cccc@{}}
\toprule
\multicolumn{10}{@{}l}{\textit{(a) Build-Time vs.\ Zero-Build Comparison on Standard Dialogue Benchmarks (Open-Weights Readers; 3-Avg = 8B, 12B, 27B)}} \\
\multirow{2}{*}{\textbf{Method}} & \textbf{Build $\downarrow$} & \multicolumn{4}{c|}{\textbf{\textsc{LongMemEval-S} (110k)}} & \multicolumn{4}{c}{\textbf{\textsc{LoCoMo} (22k, Cat 1--4)}} \\
 & \textbf{(LME-S / LoCoMo)} & 8B $\uparrow$ & 27B $\uparrow$ & 3-Avg $\uparrow$ & Tok $\downarrow$ & 8B $\uparrow$ & 27B $\uparrow$ & 3-Avg $\uparrow$ & Tok $\downarrow$ \\
\midrule
\textsc{Mem0} \citep{chhikara2025mem0}            & 50.0M / 0.4M & 38.4 & 68.2 & 48.1 & 0.7k & 37.7$^\dagger$ & 43.5$^\dagger$ & 39.7$^\dagger$ & 0.7k \\
\textsc{A-MEM} \citep{xu2025amem}                 & 35.6M / 0.5M & 49.0 & 61.2 & 54.1 & 1.7k & 67.1$^\dagger$ & 73.0$^\dagger$ & 68.8$^\dagger$ & 1.8k \\
\textsc{LightMem} \citep{fang2026lightmem}        & 32.3M / 1.0M & 67.0 & 73.0 & 66.7 & 0.8k & 68.6$^\dagger$ & 74.4$^\dagger$ & 70.9$^\dagger$ & 0.8k \\
\textsc{HippoRAG 2} \citep{gutierrez2025hipporag2}& 30.1M / 0.6M & 67.2 & 82.8 & 73.0 & 1.7k & 61.7$^\dagger$ & 74.3$^\dagger$ & 67.6$^\dagger$ & 0.9k \\
\midrule
\textsc{Raw Reranker} ($\tau{=}.05$)              & 0            & 55.0 & 58.0 & 55.9 & \textbf{0.4k} & 72.5 & 72.7 & 70.1 & \textbf{0.6k} \\
\textsc{NaiveRAG}                                 & 0            & 67.6 & 77.0 & 71.0 & 1.3k & 65.3 & 65.8 & 63.2 & 1.3k \\
\textsc{Similarity Rep.\ (FacLoc)}                & 0            & 72.0 & 83.0 & 74.8 & 1.6k & 68.1 & 68.4 & 65.8 & 1.4k \\
\textsc{Rerank-TopK}                              & 0            & \ul{74.8} & \ul{85.4} & \ul{78.1} & 1.5k & \ul{74.6} & \textbf{77.5} & \ul{73.9} & 1.3k \\
\textbf{\method{} (Ours)}                         & \textbf{0}   & \textbf{79.4} & \textbf{86.2} & \textbf{80.7} & \ul{0.6k} & \textbf{75.6} & \ul{76.9} & \textbf{74.5} & \ul{0.7k} \\
\midrule
\midrule
\multicolumn{10}{@{}l}{\textit{(b) Scaling to 1.5M Tokens (\textsc{LME-M}), Closed API (\textsc{GPT-4o-mini}), and Dense OOD (\textsc{UtilMem}, 0--100 Rubric)}} \\
\multirow{2}{*}{\textbf{Zero-Build Method}} & \textbf{\textsc{LME-S}} & \multicolumn{4}{c|}{\textbf{\textsc{LongMemEval-M} (1.5M)}} & \multicolumn{4}{c}{\textbf{\textsc{UtilMem} (OOD, 4.76 Gold Sess.)}} \\
 & \textbf{4o-m $\uparrow$} & 8B $\uparrow$ & 27B $\uparrow$ & 4o-m $\uparrow$ & Tok $\downarrow$ & 8B $\uparrow$ & 27B $\uparrow$ & 4o-m $\uparrow$ & Tok $\downarrow$ \\
\midrule
\textsc{Raw Reranker} ($\tau{=}.05$) & 57.8 & 51.0 & 55.2 & 55.4 & \textbf{0.4k} & 34.6 & 26.6 & 39.4 & \textbf{0.6k} \\
\textsc{NaiveRAG}                    & 75.2 & 60.4 & 70.8 & 68.0 & 1.3k & 33.8 & 29.7 & 41.6 & 1.6k \\
\textsc{Similarity Rep.\ (FacLoc)}   & 80.2 & 60.4 & 73.2 & 69.8 & 1.6k & 30.4 & 26.7 & 37.5 & 1.6k \\
\textsc{Rerank-TopK}                 & \ul{83.6} & \ul{68.8} & \ul{76.8} & \ul{75.0} & 1.5k & \textbf{35.7} & \ul{33.7} & \textbf{42.6} & 1.4k \\
\textbf{\method{} (Ours)}            & \textbf{85.8} & \textbf{71.6} & \textbf{79.4} & \textbf{78.0} & \ul{0.7k} & \ul{35.5} & \textbf{34.2} & \ul{42.2} & \ul{1.4k} \\
\bottomrule
\end{tabular}}
\caption{Main QA results under \textsc{DeepSeek-V4.1-Flash}: accuracy (\%) on \textsc{LME-S/M} ($n=500$) and \textsc{LoCoMo} Cat 1--4 ($n=1{,}540$), and 0--100 rubric score on \textsc{UtilMem} ($n=485$--$490$ of 500 after API moderation filters; Table~\ref{tab:utilmem_local} gives unfiltered $n=500$). \textbf{Bold}/\ul{underline}: best/2nd; Build: 8B offline tokens; Tok: mean prompt tokens; 3-Avg: 8B/12B/27B mean. $^\dagger$Build-time \textsc{LoCoMo} omits Cat-5 refusal; full tables in Appendix~\ref{app:qa_ablations}--\ref{app:profiling}.}
\label{tab:main}
\end{table*}

\section{Experiments}
\label{sec:experiments}

We evaluate three questions. First, does single-pass selection improve the accuracy--efficiency frontier up to 1.5M tokens (\textbf{RQ1}, \S\ref{sec:rq1}--\S\ref{sec:efficiency})? Second, what do pool comparison and the bidirectional head add to accuracy and order stability (\textbf{RQ2}, \S\ref{sec:rq2})? Third, how does one threshold adapt across sparse, unanswerable, and evidence-heavy queries (\textbf{RQ3}, \S\ref{sec:rq3})?

\subsection{Experimental Setup}

\paragraph{Training Data and Benchmarks.}
\method{} trains offline on the 537 synthetic dialogues from \S\ref{sec:set_calib} (Table~\ref{tab:contam}) and transfers zero-shot to four benchmarks. \textbf{\textsc{LongMemEval-S}} and \textbf{\textsc{LongMemEval-M}} \citep{wu2025longmemeval} each test 500 queries over 110k- and 1.5M-token histories. \textbf{\textsc{LoCoMo-1986}} \citep{maharana2024locomo} spans ten dialogues with answerable Category 1--4 and unanswerable Category-5 queries. Finally, \textbf{\textsc{UtilMem}} \citep{qing2026utilmem} tests evidence-heavy out-of-domain queries averaging 4.76 gold sessions across QA and turn-selection subsets.

\paragraph{Readers and Baselines.}
We serve five frozen readers at temperature $0$ via vLLM \citep{kwon2023vllm} or API: \textsc{Qwen3-8B}, \textsc{Qwen3-14B} \citep{yang2025qwen3}, \textsc{Mistral-Nemo-12B}, \textsc{Qwen3.5-27B}, and \textsc{GPT-4o-mini}. Baselines span six build-time systems in Tables~\ref{tab:main}--\ref{tab:pareto}, native 262k \textsc{FullContext}, and official \textsc{MemLoc-4B}. Over the shared front-end, we compare \textsc{NaiveRAG}, \textsc{Rerank-TopK}, \textsc{FacLoc}, \textsc{Adaptive-$k$} \citep{taguchi2025adaptivek}, \textsc{Provence} \citep{chirkova2025provence}, post-hoc rescalers, \textsc{SmartSearch}-style cutoffs, and zero-shot LLM gating. We also train four matched controls: 435M \textsc{Set-Encoder}-style, \textsc{Pointwise-1.7B}, \textsc{CausalMLP-1.7B}, and \textsc{Indep+Bidir} (Table~\ref{tab:baselines_spec}). Selectors use $B=1000$, except $B=650$ on \textsc{LoCoMo} where $B=1000$ matches recall.

\paragraph{Evaluation Protocol.}
Main QA uses unified \textsc{DeepSeek-V4.1-Flash} judging, cross-validated against archived \textsc{GPT-4o-2024-11-20} and local \textsc{Qwen3.5-27B} at 95.6\%--97.4\% agreement ($\kappa = 0.89$--$0.93$, Appendix~\ref{app:profiling}). For \textsc{UtilMem}, we also report unfiltered results under the local 27B judge. Significance uses 10,000-resample clustered bootstrap \citep{efron1993bootstrap} with Holm--Bonferroni correction \citep{holm1979simple}.

\begin{figure*}[t]
    \centering
    \includegraphics[width=\textwidth]{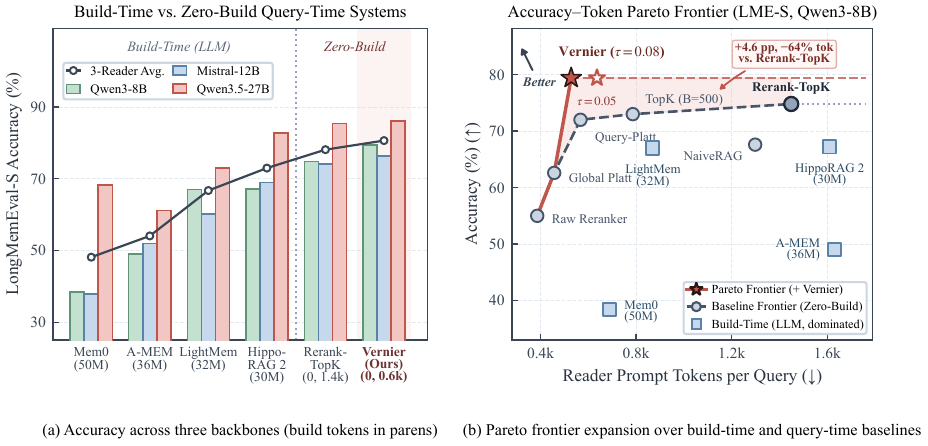}
    \caption{Accuracy staircase and Pareto frontier on 110k-token \textsc{LongMemEval-S} ($n=500$). (a) Zero-build \textsc{Rerank-TopK} outperforms build-time systems requiring 30M--50M offline tokens, and \method{} climbs higher at 0.6k prompt tokens. (b) On \textsc{Qwen3-8B}, \method{} at compact $\tau=0.08$ and default $\tau=0.05$ pushes the accuracy--token Pareto frontier up and left over all compared baselines.}
    \label{fig:pareto}
\end{figure*}

\subsection{Main Results (RQ1)}
\label{sec:rq1}

\paragraph{Build-Time vs.\ Verbatim Memory.}
Offline builders spend tens of millions of LLM tokens but drop verbatim dates and phrasing, so zero-build \textsc{Rerank-TopK} already beats every build-time baseline on \textsc{LongMemEval-S} (Table~\ref{tab:main}a, Figure~\ref{fig:pareto}a). Joint comparison lifts \method{} to \textbf{80.7\%} three-reader average at 0.6k tokens, beating \textsc{HippoRAG~2} by \textbf{+7.7\,pp} and \textsc{LightMem} by \textbf{+14.0\,pp} with zero LLM build cost. On \textsc{LoCoMo}, \method{} beats all matched build-time systems across open readers though their official prompts omit Category-5 refusal. Only \textsc{SimpleMem} edges out default \method{} on 27B at much higher token cost while collapsing on \textsc{LME-S}; disabling pruning at $\tau=0$ puts \method{} ahead.

\paragraph{Million-Token Scaling, Full Context, and Evidence-Heavy Boundaries.}
Table~\ref{tab:main}(b) scales to 1.5M-token \textsc{LongMemEval-M}, \textsc{GPT-4o-mini}, and evidence-heavy OOD \textsc{UtilMem}. Under primary seed \texttt{s42} pooled across all five readers (Table~\ref{tab:full_5reader}), \method{} beats \textsc{Rerank-TopK} by \textbf{+2.48\,pp} on \textsc{LME-S} and \textbf{+2.76\,pp} on \textsc{LME-M} (Holm $p \le 0.014$). It achieves these gains with 50\%--57\% fewer tokens (Figure~\ref{fig:pareto}b), curbing attention dilution over the same Top-32 pool. Across three symmetric seeds under identical $k_{\min}=2$ packing (Table~\ref{tab:multiseed_e2e}), every seed beats \textsc{Rerank-TopK} on 1.5M \textsc{LME-M}, averaging 71.6\% on 8B and 79.0\% on 27B at 58\% fewer tokens. On \textsc{LME-S}, every seed lifts 8B to an average of 78.3\% versus 74.8\%, while 27B averages 84.9\% versus 85.4\% at 61\% fewer tokens. Against native 262k \textsc{FullContext} on 27B \textsc{LME-S}, \method{} gains \textbf{+9.8\,pp} using \textbf{170$\times$ fewer tokens} (Holm $p=3\times 10^{-6}$). On \textsc{LoCoMo} (Table~\ref{tab:locomo_cat}), \method{} outperforms \textsc{Rerank-TopK} at 42\% fewer tokens, beats \textsc{MemLoc-4B} by \textbf{+16.8--16.9\,pp}, and lifts 8B Category-5 refusal over \textsc{FullContext} from 22.7\% to \textbf{57.4\%}.

On evidence-heavy \textsc{UtilMem}, the same $\tau=0.05$ automatically expands context roughly twofold to 1.4k tokens under identical prompts. \method{} matches full-budget \textsc{Rerank-TopK} pooled across readers while significantly beating \textsc{FacLoc}, \textsc{Raw Reranker}, and \textsc{NaiveRAG} by +2.31 to +5.85 points (Holm $p \le 0.0008$). On this 500-query QA subset, \method{} lifts mean evidence-session recall over \textsc{Rerank-TopK} from 62.3\% to 72.9\% (Table~\ref{tab:error_decomp}). However, it still trails by 1.58 points under the unfiltered local-27B judge (Holm $p=0.11$; Table~\ref{tab:utilmem_local}). Thus, broader session coverage alone does not guarantee stronger dense-synthesis QA under budget saturation.

\subsection{System Efficiency and Latency}
\label{sec:efficiency}

\begin{table}[t]
\centering
\renewcommand{\arraystretch}{0.95}
\resizebox{\columnwidth}{!}{%
\begin{tabular}{@{}l|cccc@{}}
\toprule
\textbf{System} & \textbf{Build $\downarrow$} & \textbf{Lat.\ (s) $\downarrow$} & \textbf{Tok $\downarrow$} & \textbf{Acc.\ $\uparrow$} \\
\midrule
\multicolumn{5}{@{}l}{\textit{LongMemEval-S (110k; Qwen3-8B / 27B$^\ast$)}} \\
\textsc{SimpleMem} (8B)            & 87.9M & 8.83 & 3.6k & 37.6 \\
\textsc{A-MEM} (8B)                & 35.6M & \textbf{1.09} & 1.4k & 49.0 \\
\textsc{LightMem} (8B)             & 32.3M & \ul{1.22} & \textbf{0.6k} & 67.0 \\
\textsc{HippoRAG 2} (8B)           & 30.1M & 1.57 & 1.4k & 67.2 \\
\textsc{FullContext} (27B$^\ast$)  & \textbf{0} & 50.4$^\ddagger$ & 110.4k & 76.4 \\
\textsc{Rerank-TopK} (8B/27B$^\ast$)& \textbf{0} & 1.55 & 1.4k & \ul{74.8 / 85.4} \\
\textbf{\method{} (8B / 27B$^\ast$)} & \textbf{0} & 1.55 & \textbf{0.6k} & \textbf{79.4 / 86.2} \\
\midrule
\multicolumn{5}{@{}l}{\textit{LongMemEval-M (1.5M; Qwen3-8B / 27B$^\ast$; Sel.\ Only$^\dagger$)}} \\
\textsc{Rerank-TopK} (8B/27B$^\ast$)& \textbf{0} & \textbf{0.54}$^\dagger$ & 1.6k & \ul{68.8 / 76.8} \\
\textbf{\method{} (8B / 27B$^\ast$)} & \textbf{0} & \ul{0.58}$^\dagger$ & \textbf{0.8k} & \textbf{71.6 / 79.4} \\
\midrule
\multicolumn{5}{@{}l}{\textit{LoCoMo (22k; Qwen3-8B)}} \\
\textsc{SeCom} (8B)                & 0.6M  & 0.45 & 1.1k & 51.2 \\
\textsc{MemLoc-4B}                 & 18.3m & 2.42$^\dagger$ & 3.1k & 58.8 \\
\textsc{FullContext}               & \textbf{0} & --   & 22.4k & 71.6 \\
\textsc{Rerank-TopK}               & \textbf{0} & \textbf{0.24} & 1.3k & \ul{74.6} \\
\textbf{\method{} (Ours)}          & \textbf{0} & \ul{0.28} & \textbf{0.7k} & \textbf{75.6} \\
\bottomrule
\end{tabular}}
\caption{Controlled 1$\times$H800 \texttt{bs=1} profiling, no prefix cache. \textbf{Bold}/\ul{underline}: best/2nd per block; Acc.\ in \%; Tok: 8B tokens; Build: 8B tokens except \textsc{MemLoc-4B} in minutes (m); dash: unprofiled. $^\ast$27B. $^\dagger$Selection only (45.0\,ms + 0.18\,ms DP). $^\ddagger$2$\times$H800 TP=2 (100.8\,s 1-GPU).}
\label{tab:pareto}
\end{table}

Table~\ref{tab:pareto} profiles single-H800 \texttt{bs=1} cost without prefix caching (Table~\ref{tab:latency_detail}). \method{} indexes standard histories in under 2\,s with a 436$\times$ smaller disk footprint than \textsc{MemLoc-4B}, scaling to 1.5M-token \textsc{LME-M} in 8.04\,s. Online, the 1.7B selector completes one prefill in roughly 50\,ms within 7.25\,GB VRAM. On 1.5M-token \textsc{LME-M}, full selection takes 0.58\,s versus 0.54\,s for \textsc{Rerank-TopK}, adding roughly 45\,ms while cutting prompt tokens in half. Halving prompt length offsets selector overhead on \textsc{LME-S}, matching \textsc{Rerank-TopK} end-to-end latency at 1.55\,s with minimal overhead on \textsc{LoCoMo}. Meanwhile, 8B \method{} beats 27B \textsc{FullContext} by +3.0\,pp at 32$\times$ wall-clock and 65$\times$ GPU speedups.

\section{Analysis}
\label{sec:analysis}

\begin{figure*}[t]
    \centering
    \includegraphics[width=0.92\textwidth]{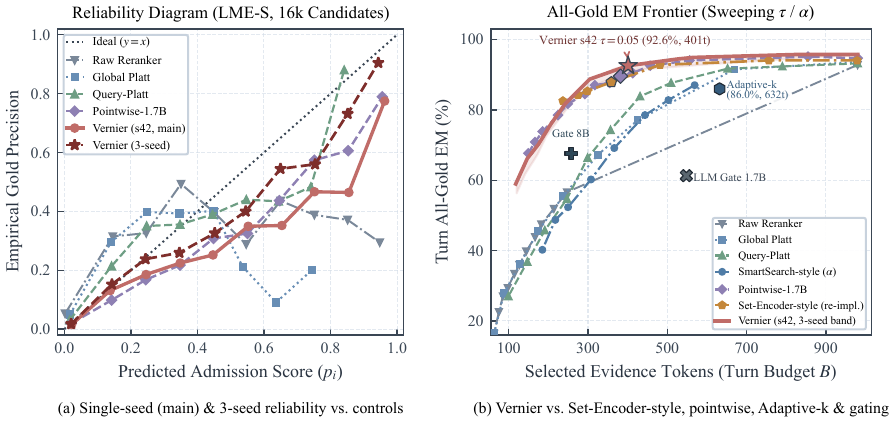}
    \caption{Admission reliability and recall--token frontier on \textsc{LME-S} ($n=16{,}000$, Table~\ref{tab:calib_full}). (a) The 3-seed \method{} ensemble aligns scores with gold precision up to $(0.94, 0.90)$, while \textsc{Pointwise-1.7B} stays overconfident and post-hoc rescalers collapse above $0.5$. (b) Sweeping $\tau$ or $\alpha$ under exact DP on 470 answerable queries, with shaded band across \texttt{s42}--\texttt{s44}, improves the recall--token trade-off.}
    \label{fig:dynamics}
\end{figure*}

\subsection{Candidate Interaction and Order Robustness (RQ2)}
\label{sec:rq2}

\begin{table}[t]
\centering
\renewcommand{\arraystretch}{0.89}
\resizebox{\columnwidth}{!}{%
\begin{tabular}{@{}l|cccc@{}}
\toprule
\multicolumn{5}{@{}l}{\textit{(a) Contextual Scoring and Evidence Admission (\textsc{LME-S}, \texttt{s42})}} \\
\textbf{Selector} & \textbf{8B $\uparrow$} & \textbf{27B $\uparrow$} & \textbf{Tok $\downarrow$} & \textbf{ms $\downarrow$} \\
\midrule
\multicolumn{5}{@{}l}{\textit{Fixed and heuristic selection}} \\
\textsc{Rerank-TopK}                        & 74.8 & 85.4 & 1,449 & -- \\
\textsc{Adaptive-$k$} (RRF, $B{=}5$)         & 72.0 & 82.4 & 953 & -- \\
\textsc{Provence} ($T{=}0.1$)                & 70.2 & 76.2 & \textbf{527} & -- \\
\textsc{SmartSearch}-style ($\alpha{=}.02$) & 73.8 & 78.8 & 679 & -- \\
\textsc{Query-Platt} ($\tau{=}0.05$)        & 72.0 & 77.8 & \ul{567} & -- \\
\midrule
\multicolumn{5}{@{}l}{\textit{Learned candidate scoring}} \\
\textsc{Set-Encoder}-style (435M)           & 78.2 & 86.0 & 595 & \textbf{36.8} \\
\textsc{Pointwise-1.7B}                     & 78.2 & 85.2 & 611 & 65.1$^\dagger$ \\
\textsc{CausalMLP-1.7B} ($\tau{=}.05$)      & 79.0 & 82.4 & \textbf{491} & -- \\
\textsc{CausalMLP-1.7B} ($\tau{=}.025$, tok-m.) & 78.6 & \textbf{87.0} & 653 & -- \\
\textsc{Indep+Bidir} (1.7B)                 & \ul{79.2} & 86.0 & 616 & 74.4 \\
\textbf{\method{} (Ours, $\tau{=}0.05$)}    & \textbf{79.4} & \ul{86.2} & 636 & \ul{53.9} \\
\midrule
\multicolumn{5}{@{}l}{\textit{(b) Core Architectural Ablations (4-Reader Pooled $\Delta$ vs.\ \method{})}} \\
\textbf{Variant} & \textbf{LME-S} & \textbf{LME-M} & \textbf{LoCoMo} & \textbf{Tok} \\
\midrule
\textit{w/o Bidir.\ Head ($\tau{=}.05$)}    & $-2.10$ & $-0.60$ & $-0.57$ & $-23\%$ \\
\textit{Similarity Rep.\ (FacLoc)}          & $-5.80$ & $-8.15$ & $-8.65$ & $+116\%$ \\
\textit{w/o Threshold ($\tau{=}0$)}         & $-1.95$ & $-2.50$ & $-0.09$ & $+110\%$ \\
\midrule
\multicolumn{5}{@{}l}{\textit{(c) Order-Reversal Stability (\textsc{LME-S} / \textsc{LME-M}, $\tau{=}0.05$)}} \\
\textbf{Model} & \textbf{Orig.\ EM} & \textbf{Rev.\ $\Delta$} & \textbf{Rev.\ $\rho$} & \textbf{Jacc.} \\
\midrule
\textsc{CausalMLP-1.7B} & 86.6 / 79.8 & $+2.76$ / $+2.34$ & .902 / .913 & 75.4 / 76.3 \\
\textbf{\method{} (Ours)} & \textbf{92.6} / \textbf{86.0} & \textbf{$+0.22$} / \textbf{$+0.21$} & \textbf{.918} / \textbf{.929} & \textbf{78.8} / \textbf{78.7} \\
\bottomrule
\end{tabular}}
\caption{RQ2 \textsc{LME-S} controls ($n=500$, \texttt{s42}), 4-reader ablations, and reversal stability (Tables~\ref{tab:perm_stress}, \ref{tab:full_5reader}). \textbf{Bold}/\ul{underline}: best/2nd in (a), better in (c); tok-m.: $\tau{=}0.025$ (Holm $p{=}1.0$). (b): pp acc./\% tok vs.\ \method{}; (c): EM (\%), shift $\Delta$, $\rho$, Jaccard (\%). $^\dagger$KV-cache; uncached 71.2/82.1\,ms.}
\label{tab:ablations}
\end{table}

\paragraph{Heuristic Cutoffs vs.\ Learned Interaction.}
Table~\ref{tab:ablations}(a) and Figure~\ref{fig:dynamics} contrast fixed, heuristic, and learned selectors on \textsc{LME-S}. Cutoff rules over isolated scores lose accuracy under compression: \textsc{Adaptive-$k$}, \textsc{Provence}, \textsc{SmartSearch}-style, and \textsc{Query-Platt} trail \method{} by 3.8--10.0\,pp across readers (Tables~\ref{tab:ans_abs}--\ref{tab:platt_qa}). Seven of eight tests stay significant at Holm $p \le 0.042$, except 27B \textsc{Adaptive-$k$} at $p=0.114$, and \textsc{Adaptive-$k$} lags further on \textsc{LME-M} and \textsc{LoCoMo} at 50\% more tokens.

Trained on identical synthetic pools, \textsc{Pointwise-1.7B} drops turn AUPRC by 6.0--10.2\,pp, runs slower even with prefix caching, and trails \method{} on \textsc{LME-S}. Faster 435M \textsc{Set-Encoder}-style and \textsc{Indep+Bidir} stay competitive: \textsc{Set-Encoder}-style reaches 72.6\% vs.\ 71.6\% on 8B \textsc{LME-M}, whereas \method{} leads 27B \textsc{LME-M} at 79.4\%, up to +3.6\,pp ahead.

\paragraph{Component Roles.}
Table~\ref{tab:ablations}(b) isolates \method{}'s components across four readers. Similarity repulsion drops QA by 5.8--8.7\,pp, while removing the cutoff at $\tau=0$ doubles context and cuts \textsc{LongMemEval} accuracy by 2.0--2.5\,pp (Holm $p=0.056$ on \textsc{LME-S}, $p=0.035$ on \textsc{LME-M}). Omitting session dates further hurts dated dialogues by up to 2.74\,pp (Table~\ref{tab:full_5reader}). A quadratic binary optimization (QUBO) probe over candidate pairs separates complementary from redundant turns at 97.3\%--98.8\% AUROC (Table~\ref{tab:pairwise_probe}), but validation selects $\beta^*=0$, indicating contextualized scores suffice for additive packing.

\paragraph{Role of the Bidirectional Set Head.}
Removing the bidirectional head (\textsc{CausalMLP-1.7B}) cuts pooled \textsc{LME-S} accuracy by 2.10\,pp at fixed $\tau=0.05$ with 23\% fewer tokens. Retuning to $\tau=0.025$ yields similar observed accuracy at matched token counts, reaching 78.6\% on 8B and 87.0\% on 27B (Holm $p=1.0$; Table~\ref{tab:ablations}a), showing that trained pruning drives the main token-matched QA gain. The bidirectional head instead delivers two architectural gains (Table~\ref{tab:ablations}c). First, it stabilizes fixed-threshold admission without retuning, lifting normal-order All-Gold EM from 86.6\% to \textbf{92.6\%} on \textsc{LME-S} and from 79.8\% to \textbf{86.0\%} on \textsc{LME-M}. Second, it yields roughly twelvefold smaller reversal-induced shifts in All-Gold EM. Reversing order shifts causal-only EM by \textbf{+2.76} and \textbf{+2.34\,pp} as early gold turns move back, whereas \method{} holds shifts to \textbf{+0.22} and \textbf{+0.21\,pp} with 78.8\% and 78.7\% Jaccard overlap.

\subsection{Automatic Budget Adaptation (RQ3)}
\label{sec:rq3}

\begin{figure}[t]
    \centering
    \includegraphics[width=0.90\columnwidth]{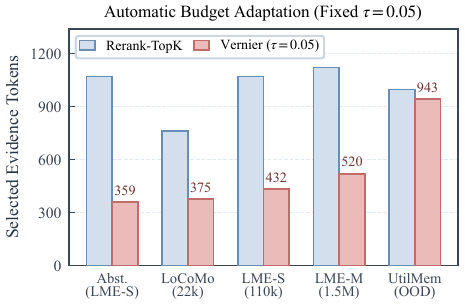}
    \caption{Automatic budget adaptation ($\tau=0.05$, $k_{\min}=2$): rendered tokens with session headers (so \textsc{TopK} can exceed turn cap $B=1000$) shrink to 359 on abstention and expand 2.6$\times$ to 943 on OOD \textsc{UtilMem}.}
    \label{fig:adaptive_budget}
\end{figure}

Figure~\ref{fig:adaptive_budget} and Table~\ref{tab:unified_budget} show one threshold adapting across sparse, unanswerable, and evidence-heavy queries. On sparse and unanswerable queries, $\tau=0.05$ prunes roughly 91\% of pools below $B$, shrinking abstention prompts to 359 tokens via $k_{\min}=2$ and lifting Category-5 refusal while preserving $B=1000$ recall. On evidence-heavy \textsc{UtilMem}, positive-utility turns saturate $B=1000$ on 94.1\% of queries, expanding context 2.6$\times$ to 943 tokens where exact DP beats score-greedy selection by \textbf{+3.08\,pp} All-Gold EM, widest at 3+ sessions. Table~\ref{tab:error_decomp} confirms 96.0\%--98.5\% Top-32 gold retention on standard benchmarks, whereas 72-token truncation and saturation constrain \textsc{UtilMem} (Table~\ref{tab:cases}).

\subsection{Category-Level Failure Diagnosis}
\label{sec:failure_diag}

Table~\ref{tab:failure_diag} disaggregates 27B \textsc{LME-S} accuracy by category. Offline builders suffer targeted omissions: \textsc{LightMem} drops to \textbf{12.5\%} on Single-Session-Assistant (SSA) turns lost in compression, while \textsc{HippoRAG~2} trails on updates and temporal reasoning. Native 262k \textsc{FullContext} degrades on multi-session (MS), temporal (TR), and abstention queries, whereas pointwise cutoffs collapse MS and TR below 33\%. Joint comparison cuts context by 56\%, leads MS and TR over \textsc{Rerank-TopK}, and matches it at 93.3\% abstention (Table~\ref{tab:ans_abs}).

\begin{table}[t]
\centering
\resizebox{\columnwidth}{!}{%
\begin{tabular}{@{}l|cccccc@{}}
\toprule
\textbf{Method (\textsc{LME-S}, 27B)} & \textbf{SSA} & \textbf{KU} & \textbf{MS} & \textbf{TR} & \textbf{Abs} & \textbf{All} \\
\midrule
\textsc{LightMem} (33.9M)     & 12.5 & 88.9 & 75.2 & 77.2 & \ul{96.7} & 73.0 \\
\textsc{HippoRAG 2} (27.6M)   & \ul{96.4} & 81.9 & 78.5 & 77.2 & \textbf{100.0} & 82.8 \\
\textsc{FullContext} (262k)   & \textbf{98.2} & 80.6 & 62.8 & 74.0 & 83.3 & 76.4 \\
\textsc{Raw Reranker}         & 94.6 & 90.3 & 25.6 & 32.3 & 93.3 & 58.0 \\
\textsc{Rerank-TopK}          & \textbf{98.2} & \textbf{95.8} & \ul{79.3} & \ul{80.3} & 93.3 & \ul{85.4} \\
\textbf{\method{} (Ours)}     & \textbf{98.2} & \ul{94.4} & \textbf{81.0} & \textbf{85.0} & 93.3 & \textbf{86.2} \\
\bottomrule
\end{tabular}}
\caption{\textsc{LME-S} 27B accuracy (\%): SSA 56, KU 72, MS 121, TR 127, Abs 30, All 500 (omitting SSU 64, SSP 30); \textbf{bold}/\ul{underline}: best/2nd. Values beside \textsc{LightMem}/\textsc{HippoRAG~2}: 27B build tokens; 262k: context.}
\label{tab:failure_diag}
\end{table}

\section{Conclusion}

Offline memory construction compacts context at heavy LLM cost and detail loss, whereas raw retrieval keeps fidelity without cross-turn selection. Through lightweight training, \method{} adds disambiguation, linking, and pruning over frozen retrieval via single-pass 1.7B prefill, bidirectional exchange, and exact packing. On \textsc{LongMemEval-S}, \method{} exceeds the strongest evaluated offline builder by 7.7\,pp across three open readers with zero LLM builds. Under primary seed \texttt{s42} across five readers, it beats full-budget reranking by 2.48\,pp on \textsc{LME-S} and 2.76\,pp on \textsc{LME-M} at half the tokens and matched \textsc{LME-S} latency. Three-seed runs confirm consistent gains on 1.5M \textsc{LME-M} and 8B readers while charting its \textsc{UtilMem} boundary. Future work explores joint front-end training and query-conditioned thresholds.

\clearpage
\section*{Limitations}

Four boundaries qualify our findings. First, because the lower \textsc{Qwen3-1.7B} backbone uses causal attention, \method{} mitigates rather than strictly eliminates order sensitivity \citep{zaheer2017deepsets}, leaving modest residual variation on evidence-heavy \textsc{UtilMem} where rank correlation sits near $\rho = 0.92$--$0.93$ (Table~\ref{tab:perm_stress}). Second, \method{} inherits the Top-32 recall ceiling of its frozen non-LLM front-end and caps turns at 72 tokens, which can truncate long multi-domain turns. It also retains modest non-LLM indexing cost alongside a roughly 50\,ms online prefill pass. Third, Eq.~\eqref{eq:knapsack} optimizes an additive utility surrogate rather than general non-additive set utility, and its selective packing gains narrow when queries saturate the token budget on evidence-heavy tasks such as \textsc{UtilMem}. Fourth, although we cross-validated three automated judges alongside targeted manual and cross-family blind audits (Appendix~\ref{app:profiling}), independent crowd-worker annotation remains future work.

\section*{Ethics Statement}

All experiments use public academic benchmarks, namely \textsc{LoCoMo}, \textsc{LongMemEval}, and \textsc{UtilMem}, together with our self-constructed synthetic training dialogues containing no real user data or personally identifiable information. Contamination checks detect no overlap between synthetic training data and any evaluation suite (Table~\ref{tab:contam}). On \textsc{UtilMem}, 10--15 of 500 queries triggered upstream HTTP 400 API safety filters in the external \textsc{DeepSeek} judge and were excluded symmetrically across all systems. Unfiltered 500-query results from our local open-weights \textsc{Qwen3.5-27B} judge appear in Table~\ref{tab:utilmem_local}.

\bibliography{custom}

\clearpage
\appendix
\raggedbottom
\setcounter{table}{0}
\renewcommand{\thetable}{A\arabic{table}}
\setcounter{figure}{0}
\renewcommand{\thefigure}{A\arabic{figure}}
\renewcommand{\topfraction}{0.90}
\renewcommand{\bottomfraction}{0.78}
\renewcommand{\textfraction}{0.10}
\renewcommand{\dbltopfraction}{0.90}
\renewcommand{\dblfloatpagefraction}{0.80}

\section{Implementation Details, Training Objective, Artifacts, and Contamination Audit}
\label{app:impl}

\subsection{Architecture, Pool Features, and Training Objective}

Table~\ref{tab:hyperparams} summarizes all architectural, optimization, and inference settings for \method{}. The frozen front-end indexes verbatim turns and session timestamps with \texttt{BAAI/bge-m3} \citep{chen2024bge} and scikit-learn TF-IDF, fuses both rankings via Reciprocal Rank Fusion with $k=60$ \citep{cormack2009rrf} to retrieve 80 turns, and trims the pool to $M=32$ candidates with \texttt{naver/provence-reranker-debertav3-v1} \citep{chirkova2025provence}. During scoring, each turn is rendered with its session date and speaker label but without numeric rank identifiers, capped at 72 tokens, and terminated by readout sentinel \texttt{</C>}; a global sufficiency sentinel \texttt{<S>} closes the prompt. Running one LM-head-free forward prefill of BF16 \texttt{Qwen/Qwen3-1.7B} \citep{yang2025qwen3} with rank-32 LoRA across all seven attention and MLP projections \citep{hu2022lora} yields candidate states $h_i^{(0)} \in \mathbb{R}^{2048}$ and global state $h_{\text{suff}}^{(0)}$.

To supply explicit pool-level context independent of prompt position, we compute a four-dimensional order-invariant feature vector $f_i \in \mathbb{R}^4$ for each candidate $c_i$ with reranker logit $s_i$, token length $\ell_i$, and session ordinal $o_i$:
\begin{equation}
\begin{aligned}
    f_i = \Bigl[ &\tfrac{s_i - \mu_s}{\max(0.5, \sigma_s)},\; \mathrm{clip}\!\bigl(\tfrac{s_i - \max_j s_j}{4}, -4, 0\bigr), \\
                 &\tfrac{\log(1+\ell_i)}{5},\; \tfrac{o_i - o_{\min}}{\max(1, o_{\max}-o_{\min})} \Bigr],
\end{aligned}
\label{eq:pool_feats}
\end{equation}
where $\mu_s$ and $\sigma_s$ are the mean and standard deviation of $\{s_j\}_{j=1}^M$. Each feature addresses a distinct selection signal. The first two entries report a turn's standardized score and margin below the top candidate, helping the selector tell sharp top hits apart from flat noisy pools. The third entry supplies log-scaled token length so the network recognizes budget cost, and the fourth supplies relative session position without exposing retrieval rank order. A two-layer GELU MLP maps $f_i$ to $\mathbb{R}^{2048}$ and adds it residually to form $\tilde{h}_i = h_i^{(0)} + \mathrm{MLP}(f_i)$. Layer-normalized states $[h_{\text{suff}}^{(0)}, \{\tilde{h}_i\}]$ pass through a two-layer Pre-LN bidirectional Transformer with 16 heads, feed-forward width 4,096, dropout $0.1$, no candidate positional embeddings, and an output LayerNorm. A shared linear head with clamped temperature $T = \mathrm{clamp}(\exp(\log T), 0.5, 3.0)$ produces logits $\theta_i = (w^\top h_i^{(2)} + b)/T$ and admission scores $p_i = \sigma(\theta_i)$, while a separate linear head on $h_{\text{suff}}^{(2)}$ outputs unanswerable-query logit $\theta_{\text{prem}}$.

During training, we randomly shuffle candidate order in every batch and optimize four complementary losses jointly. Let $y_i \in \{0,1\}$ mark gold turns, $y_i^{\text{hard}} \in \{0,1\}$ mark high-similarity contrast distractors, and $y_{\text{abs}} \in \{0,1\}$ mark unanswerable queries. First, to calibrate per-turn admission probabilities while emphasizing true evidence and deceptive distractors, the node admission loss averages weighted binary cross-entropy across valid candidates using smoothed targets $\tilde{y}_i = 0.94 y_i + 0.03$ and weights $w_i = 1 + 1.5 y_i + 0.75 y_i^{\text{hard}}$:
\begin{equation}
    \mathcal{L}_{\text{BCE}} = \frac{1}{M}\sum_{i=1}^{M} w_i\,\mathrm{BCE}\bigl(\theta_i,\, \tilde{y}_i\bigr).
    \label{eq:loss_bce}
\end{equation}
Second, to push answer-supporting turns above near-duplicate distractors within the same pool, a softplus pairwise margin term acts on answerable pools with gold set $\mathcal{P} = \{i : y_i = 1\}$ and non-gold set $\mathcal{N} = \{j : y_j = 0\}$ using margin $m = 1.5$:
\begin{equation}
\begin{aligned}
    \mathcal{L}_{\text{rank}} = \frac{1}{|\mathcal{P}||\mathcal{N}|} \sum_{i \in \mathcal{P}} \sum_{j \in \mathcal{N}} &(1 + 0.75 y_j^{\text{hard}}) \\[-2pt]
    &\cdot \log\!\bigl(1 + e^{1.5 - (\theta_i - \theta_j)}\bigr).
\end{aligned}
\label{eq:loss_rank}
\end{equation}
Third, to counter position sensitivity inherited from the causal backbone, every third micro-batch feeds a second independent shuffle $\pi_2$ of the same pool, realigns logits back to the first permutation $\pi_1$, and penalizes score drift via mean squared error:
\begin{equation}
    \mathcal{L}_{\text{perm}} = \frac{1}{M}\sum_{i=1}^{M}\Bigl(\sigma\bigl(\theta_i^{(\pi_1)}\bigr) - \sigma\bigl(\theta_i^{(\pi_2)}\bigr)\Bigr)^2.
    \label{eq:loss_perm}
\end{equation}
Fourth, an auxiliary abstention loss $\mathcal{L}_{\text{prem}} = \mathrm{BCE}(\theta_{\text{prem}}, y_{\text{abs}})$ trains the global sufficiency sentinel to detect unanswerable pools, combining Eqs.~\eqref{eq:loss_bce}--\eqref{eq:loss_perm} into the four-term joint objective of \S\ref{sec:set_calib}. We minimize this objective across seeds $\{42, 43, 44\}$ with \texttt{s42} as primary for two epochs of 289 optimizer steps at effective batch size 16, using AdamW with weight decay $0.01$, cosine decay with 8\% warmup, gradient clipping at $1.0$, learning rate $1.5\times 10^{-4}$ for LoRA, and $4.0\times 10^{-4}$ for the set head. Each seed takes 47.1--47.5 minutes, or 2,829--2,852\,s, on a single NVIDIA H800 including per-epoch synthetic validation, totaling roughly 2.4 GPU-hours across all three seeds. To favor checkpoints that retain complete evidence under tight budgets without drifting under reordering, validation selects the epoch maximizing $\text{All-Gold EM}@750 + 10 \cdot \text{RevSp}$, choosing epoch~1 for \texttt{s42} and epoch~2 for \texttt{s43} and \texttt{s44}. Before inference, we merge LoRA weights into the backbone, lock $\tau = 0.05$ once on synthetic validation, and serve open readers via deterministic vLLM \citep{kwon2023vllm} at temperature $0$. Algorithm~\ref{alg:vernier} details the full query-time procedure.

\begin{algorithm}[t]
\caption{\method{} Single-Pass Selection and Packing}
\label{alg:vernier}
\small
\begin{algorithmic}[1]
\Require Query $q$, verbatim history $\mathcal{H}$, budget $B$, threshold $\tau=0.05$, fallback $k_{\min}=2$
\Ensure Chronologically ordered evidence packet $S^*$
\State Retrieve Top-80 turns via frozen \textsc{BGE-M3} + TF-IDF RRF ($k=60$); rerank to Top-$M$ ($M=32$) via frozen \textsc{BaseProvence}, giving $\mathcal{C} = \{c_1, \dots, c_M\}$, logits $\{s_i\}$, lengths $\{\ell_i\}$.
\State Format unindexed prompt $X(q,\mathcal{C})$ by truncating each turn to 72 tokens with date/speaker headers and appending $\texttt{</C>}$ after each turn and $\texttt{<S>}$ at the end.
\State Run one LM-head-free BF16 prefill of merged \textsc{Qwen3-1.7B}+LoRA; read out $\{h_i^{(0)}\}_{i=1}^M$ and $h_{\text{suff}}^{(0)}$.
\State Compute order-invariant pool statistics $f_i$ (Eq.~\ref{eq:pool_feats}) and fuse $\tilde{h}_i \gets h_i^{(0)} + \mathrm{MLP}(f_i)$.
\State Run 2-layer position-free bidirectional encoder: $[h_{\text{suff}}^{(2)}, \{h_i^{(2)}\}] \gets \mathrm{Bidir}_2([h_{\text{suff}}^{(0)}, \{\tilde{h}_i\}])$.
\State Score $p_i \gets \sigma((w^\top h_i^{(2)} + b)/T)$ and net utility $u_i \gets p_i - \tau$ for $i \in [M]$.
\State Let $\mathcal{P}^+ = \{i \in [M] : p_i > \tau\}$.
\If{$|\mathcal{P}^+| < k_{\min}$}
    \State Greedily select up to $k_{\min}$ highest-$p_i$ turns within $B$ (keeping at least $\arg\max_i p_i$) to form $S^*$.
\Else
    \State Initialize DP table $\mathrm{dp}[0] \gets 0$, $\mathrm{dp}[1..B] \gets -\infty$.
    \For{$i \in \mathcal{P}^+$ with $\ell_i \le B$}
        \For{$b = B$ down to $\ell_i$}
            \If{$\mathrm{dp}[b - \ell_i] + u_i > \mathrm{dp}[b]$}
                \State $\mathrm{dp}[b] \gets \mathrm{dp}[b - \ell_i] + u_i$; record keep pointer.
            \EndIf
        \EndFor
    \EndFor
    \State Backtrack $S^*$ from $\arg\max_b \mathrm{dp}[b]$; if empty, set $S^* \gets \{\arg\max_i p_i\}$.
\EndIf
\State Sort $S^*$ by $(\text{session ordinal}, \text{turn position})$ and render verbatim turns with session headers.
\end{algorithmic}
\end{algorithm}

\begin{table*}[t]
\centering
\resizebox{\textwidth}{!}{%
\begin{tabular}{@{}ll|ll@{}}
\toprule
\textbf{Component / Hyperparameter} & \textbf{Value} & \textbf{Component / Hyperparameter} & \textbf{Value} \\
\midrule
Front-end Dense / Sparse & Frozen \textsc{BGE-M3} + TF-IDF, RRF $k=60$ & Optimizer \& Schedule & AdamW, weight decay $0.01$, cosine, 8\% warmup \\
Front-end Reranker & Frozen \textsc{BaseProvence}, Pool-80 $\rightarrow$ Top-32 & Learning Rates (LoRA / Head) & $1.5\times 10^{-4}$ / $4.0\times 10^{-4}$, grad clip $1.0$ \\
Selector Backbone & \textsc{Qwen3-1.7B} BF16, single prefill, 0 decode & Effective Batch / Epochs & $\text{bs}=16$ ($8\times 2$ accum.), 2 epochs, 289 steps/ep. \\
LoRA Configuration & $r=32$, $\alpha=64$, dropout $0.05$, 7 proj.\ layers & Training Loss Weights & $\mathcal{L}_{\text{BCE}} + 0.55\mathcal{L}_{\text{rank}} + 0.50\mathcal{L}_{\text{perm}} + 0.10\mathcal{L}_{\text{prem}}$ \\
Bidirectional Set Head & 2-layer Pre-LN, 16 heads, FFN 4,096, drop $0.1$ & Training Time / Seeds & 47.1--47.5\,min/seed on 1$\times$H800; seeds $\{42,43,44\}$ \\
Candidate Positional Bias & None, 72-token candidate cap, unindexed turns & Packing Threshold / Budget & Val-locked $\tau=0.05$, $k_{\min}=2$; $B=1000$ ($650$ \textsc{LoCoMo}) \\
Pool-Stats Residual MLP & 4-dim $f_i \rightarrow 512 \rightarrow 2048$ (GELU) & Reader Serving / Peak VRAM & Deterministic vLLM, temp $0$; selector peak 7.25\,GB \\
\bottomrule
\end{tabular}}
\caption{Complete architecture, optimization, and inference hyperparameters for \method{}. LoRA adapts \texttt{q/k/v/o\_proj} and \texttt{gate/up/down\_proj}. Training wall-clock time includes per-epoch synthetic validation and dual-permutation consistency passes every third micro-batch.}
\label{tab:hyperparams}
\end{table*}

\subsection{Models, Benchmarks, Licenses, and Baseline Specifications}

To support exact reproducibility, Table~\ref{tab:artifacts} records every model checkpoint, benchmark split, license, and API invocation date across our evaluation. Table~\ref{tab:baselines_spec} organizes all compared systems into three functional groups. First, build-time memory systems spend offline LLM tokens to summarize, segment, or graph-index histories before any query arrives. Second, zero-build query-time heuristics and post-hoc rescalers operate directly over the same frozen Top-32 candidate pool as \method{}, isolating whether simple cutoffs or score rescaling suffice. Third, matched learned architectural controls train on the identical synthetic split under seed \texttt{s42}, separating the effect of single-pass bidirectional set comparison from backbone scale or training data alone. All local training, profiling, and open-weights serving ran on two NVIDIA H800 80GB GPUs under BF16 precision with vLLM \citep{kwon2023vllm} at temperature $0$, \texttt{top\_p=1.0}, and \texttt{top\_k=-1}. External API calls to \textsc{GPT-4o-mini} and \textsc{DeepSeek-V4.1-Flash} ran at temperature $0$ during October--November 2026; archived \textsc{GPT-4o-2024-11-20} evaluations and the 299-item \textsc{Claude Opus 5.5} cross-family audit ran under identical frozen judging rubrics. All selector code, exact packing and evaluation scripts, trained LoRA and set-head checkpoints, the 537 synthetic dialogues, and per-query prediction logs are released publicly via the repository links on Page~1.

\begin{table*}[t]
\centering
\resizebox{\textwidth}{!}{%
\begin{tabular}{@{}llll@{}}
\toprule
\textbf{Role} & \textbf{Exact Artifact / Checkpoint Identifier} & \textbf{Scale / Split Scope} & \textbf{License / Access Notes} \\
\midrule
Dense / Reranker Front-End & \texttt{BAAI/bge-m3}; \texttt{naver/provence-reranker-debertav3-v1} & 568M / 435M, frozen & MIT / CC-BY-NC-ND-4.0 \\
Selector Backbone & \texttt{Qwen/Qwen3-1.7B} & 1.7B, LoRA $r=32$ & Apache-2.0 \\
Open-Weights Readers & \texttt{Qwen/Qwen3-8B}, \texttt{Qwen3-14B}, \texttt{Qwen3.5-27B} & 8B / 14B / 27B, temp $0$ & Apache-2.0 \\
Cross-Family Open Reader & \texttt{mistralai/Mistral-Nemo-Instruct-2407} & 12B, temp $0$ & Apache-2.0 \\
Closed API Reader & \texttt{gpt-4o-mini} (OpenAI API) & Temp $0$, Oct--Nov 2026 & Proprietary API \\
Primary / Cross Judges & \texttt{deepseek-v4.1-flash-1017} (API snapshot); \texttt{gpt-4o-2024-11-20}; \texttt{Qwen3.5-27B}; \textsc{Opus 5.5} & Temp $0$, unified rubrics & API / Apache-2.0 \\
\midrule
\textsc{LongMemEval-S / M} & \texttt{longmemeval\_s\_cleaned.json} / \texttt{m\_cleaned.json} \citep{wu2025longmemeval} & 500 queries each, 110k / 1.5M tok & MIT \\
\textsc{LoCoMo-1986} & \texttt{locomo10.json} \citep{maharana2024locomo} & 10 dialogues, 1,540 Cat 1--4 + 446 Cat-5 & CC-BY-NC-4.0 \\
\textsc{UtilMem} & \texttt{utilmem\_multi\_domain\_eval\_strong.json} \citep{qing2026utilmem} & 1,717 turn-sel.; 500 stratified QA & Research release \\
Synthetic Training Corpus & Self-constructed via \texttt{deepseek-v4.1-flash-1017} & 537 histories, 4,622 train / 766 val & Zero real user PII \\
\bottomrule
\end{tabular}}
\caption{Exact model checkpoints, public benchmarks, licenses, and API provenance across all experiments. Open-weights inference ran deterministically via vLLM on 2$\times$NVIDIA H800 80GB GPUs; API calls occurred after Oct~17, 2026 (logged snapshot \texttt{deepseek-v4.1-flash-1017}). The \textsc{Set-Encoder}-style control initialized from \textsc{BaseProvence} is used solely for academic evaluation; no derivative weights are distributed.}
\label{tab:artifacts}
\end{table*}

\begin{table*}[t]
\centering
\resizebox{\textwidth}{!}{%
\begin{tabular}{@{}lll@{}}
\toprule
\textbf{Family / Baseline} & \textbf{Exact Configuration and Hyperparameters} & \textbf{Protocol Scope} \\
\midrule
\multicolumn{3}{@{}l}{\textit{Build-Time Memory Systems and Long-Context / Iterative Controls}} \\
\textsc{Mem0}, \textsc{A-MEM}, \textsc{LightMem} & Official pipelines with matched 8B/12B/27B backbones; default extraction/update settings & \textsc{LME-S}, \textsc{LoCoMo} Cat 1--4 \\
\textsc{HippoRAG 2}, \textsc{SeCom}, \textsc{SimpleMem} & Official repos with matched open readers; offline graph/segment/summary construction & \textsc{LME-S}, \textsc{LoCoMo} Cat 1--4 \\
\textsc{MemLoc-4B} \citep{wang2026memloc} & Official \texttt{gpt-4o-mini} Stage 1--2 extraction (18.3\,min) + official 4B multi-step locator & \textsc{LoCoMo} official \& sym.\ prompt \\
\textsc{FullContext} & Full verbatim history with symmetric session headers; 8B YaRN 131k, 27B native 262k & \textsc{LME-S}, \textsc{LoCoMo} \\
\midrule
\multicolumn{3}{@{}l}{\textit{Zero-Build Query-Time Heuristics and Post-Hoc Rescalers (Shared Frozen Front-End, $M=32$)}} \\
\textsc{NaiveRAG} / \textsc{Rerank-TopK} & Stage-1a RRF or Stage-1b \textsc{BaseProvence} greedy packing up to $B$ ($1000$; $650$ \textsc{LoCoMo}) & All 4 benchmarks, 5 readers \\
\textsc{Similarity Rep.\ (FacLoc)} & Greedy submodular facility location over \textsc{BGE-M3} cosines with $\ell_i^{0.25}$ length damping & All 4 benchmarks, 5 readers \\
\textsc{Raw Reranker} ($\tau=0.05$) & Exact knapsack DP over raw \textsc{BaseProvence} $\sigma(s_i) - 0.05$ with $k_{\min}=2$ fallback & All 4 benchmarks, 5 readers \\
\textsc{Global Platt} / \textsc{Isotonic} / \textsc{Query-Platt} & Fitted strictly on synthetic val (766 queries); \textsc{Query-Platt} uses $[s_i, s_i-\max_j s_j, z_i, 1]$ & All 4 benchmarks; \textsc{LME-S} QA \\
\textsc{Adaptive-$k$} \citep{taguchi2025adaptivek} & Largest score-gap cutoff on RRF scores with buffer $B_{\text{buf}}=5$, capped by budget $B$ & \textsc{LME-S/M}, \textsc{LoCoMo} \\
\textsc{Provence} \citep{chirkova2025provence} & Official sentence-level pruning over Top-32 at thresholds $T \in \{0.1, 0.5\}$ & \textsc{LME-S} QA \\
\textsc{SmartSearch}-style \citep{derehag2026smartsearch} & Gap-ratio cutoff at validation-matched $\alpha=0.02$ and generous $\alpha=0.01$, $k_{\min}=2$ & All 4 benchmarks; \textsc{LME-S} QA \\
\textsc{LLM Recall Gating} & Zero-shot \textsc{Qwen3-1.7B} / \textsc{Qwen3-8B} via vLLM (temp $0$, max 128 output tokens) & \textsc{LME-S} QA \& latency \\
\midrule
\multicolumn{3}{@{}l}{\textit{Matched Learned Architectural Controls (Identical Synthetic Training Split and Seed \texttt{s42})}} \\
\textsc{Set-Encoder}-style (435M) & Ungated \textsc{DeBERTa-v3-large} initialized from \textsc{BaseProvence}; inter-turn attn.\ from layer 18, 3 ep. & All 4 benchmarks; \textsc{LME-S/M} QA \\
\textsc{Pointwise-1.7B} & Identical \textsc{Qwen3-1.7B}+LoRA ($r=32$, 2 ep.) scoring each $(q, c_i)$ pair independently & All 4 benchmarks; \textsc{LME-S} QA \\
\textsc{CausalMLP-1.7B} & Identical joint prefill and losses as \method{}, replacing 2-layer bidirectional head with 2-layer MLP & All 4 benchmarks, 5 readers \\
\textsc{Indep+Bidir} (1.7B) & Independent $(q, c_i)$ \textsc{Qwen3-1.7B}+LoRA encoding followed by identical 2-layer bidirectional head & All 4 benchmarks; \textsc{LME-S/M} QA \\
\bottomrule
\end{tabular}}
\caption{Implementation and hyperparameter specifications for all compared build-time baselines, query-time heuristics, and matched learned controls. All zero-build query-time selectors share the identical frozen \textsc{BGE-M3}+TF-IDF+\textsc{BaseProvence} Top-32 candidate pool and symmetric reader prompt formatting.}
\label{tab:baselines_spec}
\end{table*}

\subsection{Capability-Driven Synthetic Corpus and Contamination Audit}

Rather than mimicking the domains, personas, or lengths of \textsc{LoCoMo}, \textsc{LongMemEval}, or \textsc{UtilMem}, we construct a compact synthetic corpus offline with \textsc{DeepSeek-V4.1-Flash} solely to supply QA-grounded supervision for answer-supporting evidence and reader distractor resistance that standard single-document relevance datasets lack. Using deterministic storyline blueprints spanning 60 persona names, 20 four-topic arcs, and 2023--2025 calendars, we generate 537 multi-session dialogues balanced across six everyday domains and partition them strictly by dialogue history into 459 training and 78 validation histories with zero shared dialogues or speakers (Table~\ref{tab:synth_stats}). Every dialogue is routed through the identical frozen Top-32 front-end to expose authentic candidate traps and score collapse while turning histories of arbitrary length into a domain-agnostic 32-turn comparison task.

Each corpus design rule targets one specific capability required by \method{}. First, to teach \emph{cross-turn disambiguation}, evolving story arcs embed state updates, relative-time anchors, and dated sequences so the selector learns to keep current or target-date facts while rejecting near-duplicate superseded plans. Second, to teach \emph{complementary multi-turn retention}, cross-chapter bridges, scattered aggregations, and coreference queries carry multi-turn gold sets averaging 1.79 turns, up to 8, so the selector preserves linked turns together rather than discarding shared-entity evidence. Third, to teach \emph{noise filtering and calibration}, 10.0\% of queries pose false premises with empty gold sets so the selector learns to suppress irrelevant or empty pools below a fixed threshold. Fourth, to \emph{block surface shortcuts}, long-horizon histories intentionally mix 12--42-word colloquial exchanges with 110--210-word structured turns so length cannot predict relevance, while training strips retrieval rank IDs and shuffles candidate order every batch to prevent positional cues.

Table~\ref{tab:contam} audits potential data leakage between the 4,622 synthetic training queries and all four evaluation benchmarks. Exact question matching, 8-gram overlap, and 95th-percentile 5-gram Jaccard similarity check for surface phrasing overlap, while maximum \textsc{BGE-M3} dense cosine similarity checks for paraphrased turn leakage; every metric confirms zero contamination. Beyond zero overlap, three empirical checks mitigate concerns about synthetic-style shortcutting. First, \textsc{Pointwise-1.7B} trains the identical \textsc{Qwen3-1.7B} backbone on the exact same synthetic pools but drops 6.0--10.2\,pp turn AUPRC and trails in QA (Tables~\ref{tab:ablations} and~\ref{tab:calib_full}), while \textsc{CausalMLP-1.7B} suffers +2.34--2.76\,pp reversal drift (Table~\ref{tab:perm_stress}), isolating gains to bidirectional set comparison. Second, although synthetic histories span only 1.2k--8.4k words, \method{} transfers zero-shot to 1.5M-token \textsc{LongMemEval-M} and out-of-domain \textsc{UtilMem} financial, clinical, and study dialogues, where the same fixed threshold automatically expands context 2.6$\times$. Third, all main conclusions replicate under archived \textsc{GPT-4o} and local \textsc{Qwen3.5-27B} judges independent of \textsc{DeepSeek-V4.1-Flash} (Appendix~\ref{app:profiling}).

\begin{table*}[t]
\centering
\resizebox{\textwidth}{!}{%
\begin{tabular}{@{}lcccccc@{}}
\toprule
\textbf{Corpus Subset / Split} & \textbf{Histories} & \textbf{Sessions / Hist.} & \textbf{Mean Turns / Words} & \textbf{Queries} & \textbf{Top-32 Cand.\ (\% Pos.)} & \textbf{Abstention Queries} \\
\midrule
Long-Horizon Blueprints & 217 & 32 (4 chapters) & 267.1 turns / $\sim$8.4k words & -- & -- & -- \\
Medium-Horizon Blueprints & 320 & 8 & 57.2 turns / $\sim$1.2k words & -- & -- & -- \\
\midrule
Train Split (strict history split) & 459 & 8--32 & 142.0 turns overall & 4,622 & 147,904 (4.64\%) & 460 (10.0\%) \\
Validation Split (strict history split) & 78 & 8--32 & 142.0 turns overall & 766 & 24,512 (4.77\%) & 79 (10.3\%) \\
\textbf{Total Synthetic Corpus} & \textbf{537} & \textbf{8--32} & \textbf{76,254 total turns} & \textbf{5,388} & \textbf{172,416 (4.66\%)} & \textbf{539 (10.0\%)} \\
\bottomrule
\end{tabular}}
\caption{Statistics of the zero-overlap synthetic training corpus generated with \textsc{DeepSeek-V4.1-Flash}. Queries are balanced across six domains (friends chat, user--assistant logs, team syncs, advisor/clinician follow-ups, research collaboration, and family groups; 881--916 queries per domain). The 4,849 answerable queries average 1.79 gold turns, ranging from 1 to 8; dashes denote summary rows combined before train/val partitioning.}
\label{tab:synth_stats}

\vspace{7pt}
\resizebox{\textwidth}{!}{%
\begin{tabular}{@{}lccccccc@{}}
\toprule
\textbf{Evaluation Benchmark} & \textbf{Eval Queries} & \textbf{Avg.\ Hist.\ Tok} & \textbf{Exact Q Match} & \textbf{Q 8-gram (\%)} & \textbf{Q 5-gram Jacc.\ (p95)} & \textbf{Cand.\ $\cos{>}0.90$} & \textbf{Cand.\ p99 $\cos$} \\
\midrule
\textsc{LoCoMo-1986} (\texttt{locomo10.json}) & 1,986 (1,540+446) & $\sim$22k & 0 & 0.00\% & 0.0000 & 0.00\% & 0.7146 \\
\textsc{LongMemEval-S} (\texttt{cleaned.json}) & 500 & $\sim$110k & 0 & 0.00\% & 0.0358 & 0.00\% & 0.7257 \\
\textsc{LongMemEval-M} (\texttt{cleaned.json}) & 500 & $\sim$1.5M & 0 & 0.00\% & 0.0358 & 0.00\% & 0.7257 \\
\textsc{UtilMem} (\texttt{eval\_strong.json}) & 1,717 (500 E2E) & Multi-dom.\ & 0 & 0.00\% & 0.0000 & 0.00\% & 0.7369 \\
\bottomrule
\end{tabular}}
\caption{Benchmark statistics and contamination audit of the 4,622-query synthetic training split against all four evaluation suites. For \textsc{LoCoMo}, 1,540+446 denotes Category 1--4 answerable plus Category-5 adversarial queries; for \textsc{UtilMem}, 1,717 queries cover turn-selection evaluation across five domains (\texttt{edgar\_10k}, \texttt{finance}, \texttt{fitness}, \texttt{mental\_health}, \texttt{studychat}) and 500 stratified queries (100 per domain) form the end-to-end QA subset. Q 8-gram and Q 5-gram Jacc.\ (p95) measure lexical question overlap; Cand.\ $\cos>0.90$ and p99 $\cos$ measure maximum \textsc{BGE-M3} cosine similarity between synthetic training turns and benchmark candidates.}
\label{tab:contam}
\end{table*}
\FloatBarrier

\section{Extended Turn-Selection Calibration, Multi-Seed Stability, and Pairwise Probe}
\label{app:calib}

\subsection{Calibration, Post-Hoc Rescaling, and Three-Seed Stability}

A central diagnostic question of this paper is whether fixed-threshold pruning fails merely because raw reranker logits are miscalibrated, or because pointwise scoring lacks cross-turn comparison. To disentangle both effects, Table~\ref{tab:calib_full} evaluates raw reranker scores, three post-hoc rescalers fitted strictly on the 766-query synthetic validation split, and all neural controls across ten turn-selection metrics. For calibration, 15-bin Expected Calibration Error \citep{guo2017calibration} measures the gap between predicted admission scores and empirical gold frequency, while Brier score \citep{brier1950verification} and negative log-likelihood penalize inaccurate or overconfident probabilities. For discrimination, AUROC and AUPRC measure threshold-free separation of gold turns from distractors. For downstream packing, $\text{Gold}<0.05$ tracks severe score collapse where true evidence falls below our fixed cutoff, Turn All-Gold EM measures whether every required gold turn enters the final packed packet, $\text{Bind}\%$ records how often selection saturates budget $B$, and $\text{RevSp}$ measures Spearman rank stability when the 32-turn candidate order is reversed.

Table~\ref{tab:calib_full} reveals why post-hoc score rescaling alone cannot solve evidence selection. \textsc{Global Platt} and \textsc{Isotonic} \citep{zadrozny2002isotonic} are monotonic mappings of raw reranker logits, so they reduce ECE without improving ranking order, leaving 44.0\%--56.3\% of \textsc{LongMemEval} gold turns collapsed below $0.05$. Adding pool-relative statistics in four-parameter \textsc{Query-Platt} over $[s_i, s_i - \max_j s_j, z_i, 1]$ lifts \textsc{LME-S} Turn All-Gold EM from 47.45\% to 74.26\%, confirming that relative pool context helps substantially. However, because \textsc{Query-Platt} still cannot read candidate text jointly, it remains 18.29\,pp behind \method{}'s 92.55\% on \textsc{LME-S} and collapses further on multi-session \textsc{LME-M} and \textsc{UtilMem}. We also checked whether an extra post-hoc temperature on \method{} improves calibration beyond our jointly learned $T=1.018$. Fitting $T_{\text{post}}=0.877$ on validation mildly lowers ECE on \textsc{LoCoMo} from $0.0510$ to $0.0420$ and on \textsc{LME-S} from $0.0249$ to $0.0221$ at identical AUROC and AUPRC, but sharper scores push borderline gold turns below $\tau=0.05$ and drop \textsc{LME-S} Turn All-Gold EM from $92.55\%$ to $90.64\%$; we therefore retain the jointly learned temperature.

To isolate true initialization variance under one deployment rule, Table~\ref{tab:multiseed_e2e} evaluates three independent seeds \texttt{s42}, \texttt{s43}, and \texttt{s44} using identical $\tau=0.05$ and $k_{\min}=2$ packing ($0.0\%$ of queries below two turns) and identical synthetic-validation checkpoint selection under the deployed rule. Across 1.5M-token \textsc{LongMemEval-M}, every seed outperforms full-budget \textsc{Rerank-TopK} by $+2.6$ to $+3.0$\,pp on 8B at $71.60 \pm 0.20\%$ and by $+1.8$ to $+2.6$\,pp on 27B at $79.00 \pm 0.40\%$ while cutting prompt tokens by 58\%. On \textsc{LongMemEval-S} and \textsc{LoCoMo} All-5, every seed likewise leads \textsc{Rerank-TopK} on \textsc{Qwen3-8B}, averaging $78.27 \pm 1.10\%$ versus $74.80\%$ on \textsc{LME-S} and $71.90 \pm 0.41\%$ versus $69.44\%$ on \textsc{LoCoMo}, with \texttt{s43} and \texttt{s44} lifting 8B Category-5 refusal above 60\%. On 27B \textsc{LME-S}, \texttt{s43} and \texttt{s44} average $84.2\%$ versus $86.2\%$ for \texttt{s42} and $85.4\%$ for \textsc{Rerank-TopK}, yielding a 3-seed mean of $84.87 \pm 1.15\%$ at 61\% fewer tokens. Turn-selection diagnostics trace this gap to cross-seed calibration sharpness rather than ranking quality: all three seeds achieve nearly identical threshold-free ranking at $\tau=0$ ($95.74\%$, $95.32\%$, and $95.32\%$ on \textsc{LME-S}; $88.30\%$ across all three seeds on \textsc{LME-M}), whereas \texttt{s43} and \texttt{s44} converge to sharper Epoch-2 probability tails that pack 15\%--21\% fewer prompt tokens than \texttt{s42}.

\begin{table*}[t]
\centering
\renewcommand{\arraystretch}{0.91}
\resizebox{\textwidth}{!}{%
\begin{tabular}{@{}llccccccccc@{}}
\toprule
\textbf{Bench. ($B$)} & \textbf{Scorer} & \textbf{ECE $\downarrow$} & \textbf{Brier $\downarrow$} & \textbf{NLL $\downarrow$} & \textbf{AUROC $\uparrow$} & \textbf{AUPRC $\uparrow$} & \textbf{Gold$<$.05 $\downarrow$} & \textbf{EM $\uparrow$} & \textbf{Tok (Bind\%)} & \textbf{RevSp $\uparrow$} \\
\midrule
\multirow{9}{*}{\shortstack[l]{\textsc{LoCoMo}\\($B{=}650$)}}
 & \textsc{BaseProv Raw}      & 0.0566 & 0.0492 & 0.1716 & 89.88 & 33.83 & 17.2\% & 68.29 & 244 (11.7\%) & -- \\
 & \textsc{Global Platt}      & 0.0480 & 0.0295 & 0.1252 & 89.88 & 33.83 & 6.8\%  & 71.22 & 375 (28.1\%) & -- \\
 & \textsc{Isotonic}          & 0.0479 & 0.0298 & 0.1287 & 89.79 & 31.11 & 9.6\%  & 70.38 & 326 (21.0\%) & -- \\
 & \textsc{Query-Platt}       & \textbf{0.0128} & 0.0222 & \textbf{0.0822} & 95.16 & 50.79 & 9.1\%  & 70.57 & 212 (1.5\%)  & -- \\
 & \textsc{Set-Encoder-435M}  & 0.0613 & 0.0361 & 0.1324 & 95.57 & 63.93 & 4.5\%  & 73.96 & 277 (--)     & \textbf{0.9994} \\
 & \textsc{Pointwise-1.7B}    & 0.1188 & 0.0591 & 0.2109 & 95.17 & 59.17 & \textbf{1.6\%} & 74.87 & 420 (33.1\%) & -- \\
 & \textsc{Indep+Bidir (1.7B)}& 0.0659 & 0.0328 & 0.1247 & 96.50 & 65.67 & 2.5\%  & 74.41 & 336 (13.0\%) & -- \\
 & \textsc{CausalMLP-1.7B}    & 0.0329 & \textbf{0.0202} & 0.0830 & 96.83 & 67.90 & 5.1\%  & 72.53 & 222 (3.2\%)  & 0.8732 \\
 & \textbf{\method{} (s42 / 3-seed)} & 0.0510 / 0.0400 & 0.0273 / 0.0218 & 0.1077 / 0.0881 & \textbf{97.20} / 97.14 & 68.67 / \textbf{69.34} & 2.1\% / 3.2\% & \textbf{75.07} / 74.22 & 314 (8.0\%) & 0.8992 \\
\midrule
\multirow{9}{*}{\shortstack[l]{\textsc{LME-S}\\($B{=}1000$)}}
 & \textsc{BaseProv Raw}      & 0.0633 & 0.0679 & 0.3934 & 80.81 & 24.91 & 67.3\% & 47.45 & 201 (1.0\%)  & -- \\
 & \textsc{Global Platt}      & 0.0405 & 0.0603 & 0.2321 & 80.81 & 24.91 & 51.1\% & 58.51 & 269 (2.8\%)  & -- \\
 & \textsc{Isotonic}          & 0.0424 & 0.0614 & 0.2404 & 80.75 & 24.44 & 56.3\% & 54.68 & 242 (2.2\%)  & -- \\
 & \textsc{Query-Platt}       & 0.0229 & 0.0530 & 0.1871 & 87.61 & 36.95 & 23.8\% & 74.26 & 357 (5.4\%)  & -- \\
 & \textsc{Set-Encoder-435M}  & 0.0259 & 0.0399 & 0.1531 & 90.86 & 63.89 & 18.6\% & 87.87 & 359 (--)     & \textbf{0.9993} \\
 & \textsc{Pointwise-1.7B}    & 0.0219 & 0.0447 & 0.1686 & 87.83 & 55.10 & 18.8\% & 89.57 & 382 (8.8\%)  & -- \\
 & \textsc{Indep+Bidir (1.7B)}& 0.0226 & 0.0391 & 0.1478 & 90.52 & 64.36 & 13.8\% & 90.00 & 382 (7.4\%)  & -- \\
 & \textsc{CausalMLP-1.7B}    & \textbf{0.0058} & \textbf{0.0369} & 0.1419 & 89.60 & 64.80 & 20.7\% & 86.81 & 273 (4.4\%)  & 0.8993 \\
 & \textbf{\method{} (s42 / 3-seed)} & 0.0249 / 0.0116 & 0.0416 / 0.0377 & 0.1556 / \textbf{0.1409} & \textbf{91.52} / 91.07 & 65.24 / \textbf{65.32} & \textbf{12.3\%} / 15.8\% & \textbf{92.55} / 91.28 & 402 (7.6\%) & 0.9090 \\
\midrule
\multirow{9}{*}{\shortstack[l]{\textsc{LME-M}\\($B{=}1000$)}}
 & \textsc{BaseProv Raw}      & 0.0529 & 0.0595 & 0.3131 & 81.42 & 22.88 & 62.4\% & 45.74 & 225 (2.6\%)  & -- \\
 & \textsc{Global Platt}      & 0.0263 & 0.0520 & 0.1971 & 81.42 & 22.88 & 44.0\% & 57.02 & 317 (6.4\%)  & -- \\
 & \textsc{Isotonic}          & 0.0278 & 0.0530 & 0.2054 & 81.31 & 22.42 & 50.2\% & 53.19 & 280 (5.4\%)  & -- \\
 & \textsc{Query-Platt}       & 0.0151 & 0.0466 & 0.1651 & 87.94 & 35.60 & 20.4\% & 69.57 & 379 (5.2\%)  & -- \\
 & \textsc{Set-Encoder-435M}  & 0.0402 & 0.0380 & 0.1438 & 92.19 & \textbf{64.23} & 12.6\% & 82.13 & 402 (--) & \textbf{0.9995} \\
 & \textsc{Pointwise-1.7B}    & 0.0494 & 0.0459 & 0.1673 & 90.48 & 53.90 & 10.9\% & 84.47 & 466 (13.8\%) & -- \\
 & \textsc{Indep+Bidir (1.7B)}& 0.0402 & 0.0389 & 0.1435 & 92.68 & 62.84 & 8.1\%  & 83.62 & 444 (8.8\%)  & -- \\
 & \textsc{CausalMLP-1.7B}    & \textbf{0.0136} & \textbf{0.0332} & \textbf{0.1253} & 92.38 & 64.07 & 14.8\% & 80.21 & 309 (5.4\%)  & 0.9099 \\
 & \textbf{\method{} (s42 / 3-seed)} & 0.0444 / 0.0257 & 0.0426 / 0.0356 & 0.1571 / 0.1303 & \textbf{93.35} / 93.09 & 63.17 / 63.76 & \textbf{6.8\%} / 9.5\% & \textbf{85.96} / 84.47 & 465 (9.2\%) & 0.9196 \\
\midrule
\multirow{9}{*}{\shortstack[l]{\textsc{UtilMem}\\($n{=}1{,}717$)}}
 & \textsc{BaseProv Raw}      & 0.4232 & 0.4231 & 3.3356 & 66.66 & 59.73 & 99.9\% & 0.00  & 410 (0.0\%)  & -- \\
 & \textsc{Global Platt}      & 0.4061 & 0.4072 & 1.6980 & 66.66 & 59.73 & 98.8\% & 0.00  & 417 (0.7\%)  & -- \\
 & \textsc{Isotonic}          & 0.4022 & 0.4041 & 1.6155 & 66.01 & 56.68 & 99.6\% & 0.00  & 411 (0.1\%)  & -- \\
 & \textsc{Query-Platt}       & 0.3343 & 0.3339 & 1.1061 & 70.72 & 64.36 & 36.2\% & 15.14 & 957 (93.4\%) & -- \\
 & \textsc{Set-Encoder-435M}  & 0.3126 & 0.3212 & 1.1360 & 71.74 & 64.51 & 42.7\% & 17.76 & 934 (--)     & \textbf{0.9998} \\
 & \textsc{Pointwise-1.7B}    & 0.3134 & 0.3192 & 1.0231 & 73.61 & 64.46 & 27.1\% & 25.28 & 933 (88.4\%) & -- \\
 & \textsc{Indep+Bidir (1.7B)}& 0.2940 & 0.3054 & 1.0481 & 75.67 & 67.94 & 37.4\% & 23.53 & 916 (83.5\%) & -- \\
 & \textsc{CausalMLP-1.7B}    & 0.3260 & 0.3176 & 1.0691 & 79.04 & 71.75 & 35.4\% & 23.12 & 912 (83.3\%) & 0.9134 \\
 & \textbf{\method{} (s42 / 3-seed)} & \textbf{0.2827} / 0.2994 & \textbf{0.2918} / 0.2999 & \textbf{0.9239} / 0.9897 & 78.22 / \textbf{79.48} & 70.41 / \textbf{71.93} & \textbf{18.9\%} / 31.0\% & \textbf{26.67} / 23.47 & 964 (94.1\%) & 0.9267 \\
\bottomrule
\end{tabular}}
\caption{Reconciled turn-selection calibration, discrimination, exact-knapsack Turn All-Gold EM, and reverse-order Spearman rank stability (\textsc{RevSp}) across all four benchmarks at $\tau=0.05$ and $k_{\min}=2$. \textbf{Bold} marks the best value per benchmark block. AUROC, AUPRC, and EM are percentages; Gold$<$.05 is the fraction of gold turns scored below $0.05$; Tok (Bind\%) reports mean selected evidence tokens and percentage of queries hitting budget $B$; 3-seed averages admission scores across \texttt{s42}--\texttt{s44}. On \textsc{UtilMem}, upstream annotations specify gold sessions rather than turns, so EM measures all-gold-session coverage and turn-level calibration uses a session-derived positive turn proxy (Appendix~\ref{app:prompts_cases}). Dashes denote inherently order-invariant scorers ($\text{RevSp}=1.0$ by construction) or unlogged binding rate for \textsc{Set-Encoder-435M}. At $\tau=0$, 3-seed \method{} reaches $75.80 \pm 0.14\%$ on \textsc{LoCoMo}, $95.53 \pm 0.21\%$ on \textsc{LME-S}, $88.23 \pm 0.12\%$ on \textsc{LME-M}, and $33.45 \pm 1.32\%$ on \textsc{UtilMem}.}
\label{tab:calib_full}

\vspace{7pt}
\resizebox{\textwidth}{!}{%
\begin{tabular}{@{}ll|c|ccc|cc@{}}
\toprule
\textbf{Benchmark} & \textbf{Reader} & \textbf{\textsc{TopK} (pTok)} & \textbf{\texttt{s42} (pTok)} & \textbf{\texttt{s43} (pTok)} & \textbf{\texttt{s44} (pTok)} & \textbf{3-Seed Mean $\pm$ SD} & \textbf{Mean pTok} \\
\midrule
\multirow{2}{*}{\textsc{LongMemEval-S} ($n{=}500$)} & \textsc{Qwen3-8B} & 74.8 (1,449) & \textbf{79.4} (636) & 78.2 (510) & 77.2 (547) & \textbf{78.27 $\pm$ 1.10} & 564 ($-61\%$) \\
 & \textsc{Qwen3.5-27B} & 85.4 (1,478) & \textbf{86.2} (648) & 84.2 (518) & 84.2 (557) & 84.87 $\pm$ 1.15 & 574 ($-61\%$) \\
\midrule
\multirow{2}{*}{\textsc{LongMemEval-M} ($n{=}500$)} & \textsc{Qwen3-8B} & 68.8 (1,563) & 71.6 (769) & \textbf{71.8} (574) & 71.4 (620) & \textbf{71.60 $\pm$ 0.20} & 654 ($-58\%$) \\
 & \textsc{Qwen3.5-27B} & 76.8 (1,592) & \textbf{79.4} (783) & 79.0 (583) & 78.6 (631) & \textbf{79.00 $\pm$ 0.40} & 665 ($-58\%$) \\
\midrule
\multirow{2}{*}{\textsc{LoCoMo} All-5 ($n{=}1{,}986$)} & \textsc{Qwen3-8B} & 69.4 (1,262) & 71.5 (725) & 71.9 (569) & \textbf{72.3} (633) & \textbf{71.90 $\pm$ 0.41} & 643 ($-49\%$) \\
 & \textsc{Qwen3.5-27B} & 81.2 (1,295) & \textbf{80.3} (744) & 79.0 (584) & 79.5 (649) & 79.57 $\pm$ 0.66 & 659 ($-49\%$) \\
\bottomrule
\end{tabular}}
\caption{Symmetric end-to-end \textsc{DeepSeek} QA accuracy (\%) and mean reader prompt tokens (pTok) across three independent training seeds of \method{} under the identical deployed rule ($\tau=0.05$, $k_{\min}=2$, $0.0\%$ of queries below two turns) versus full-budget \textsc{Rerank-TopK} (\textsc{TopK}). \textbf{Bold} marks the highest \method{} seed per row and 3-seed means exceeding \textsc{TopK}; SD denotes sample standard deviation across the three seeds. On \textsc{LoCoMo} Cat 1--4 ($n=1{,}540$), 3-seed \method{} averages $75.43 \pm 0.44\%$ on 8B (vs.\ $74.61\%$ \textsc{TopK}) and $76.17 \pm 0.72\%$ on 27B (vs.\ $77.53\%$ \textsc{TopK}).}
\label{tab:multiseed_e2e}
\end{table*}

\subsection{Order-Permutation Stress Test, Budget Unification, and Pairwise QUBO Probe}

\begin{figure*}[t]
    \centering
    \includegraphics[width=0.98\textwidth]{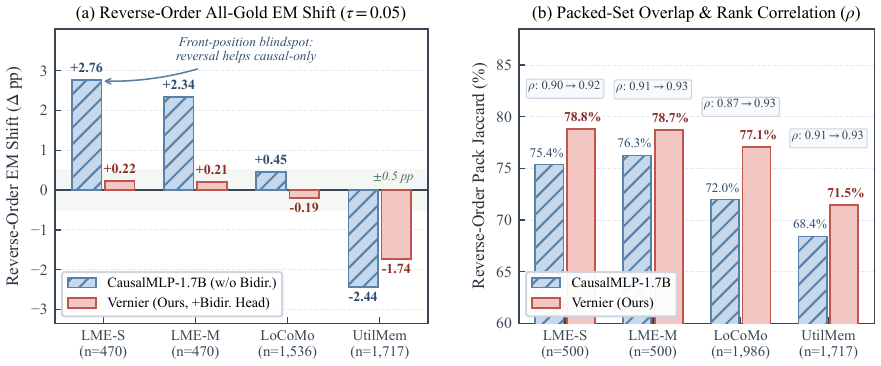}
    \caption{Candidate-order permutation stress test comparing causal-only \textsc{CausalMLP-1.7B} against \method{} at $\tau=0.05$ and $k_{\min}=2$ (full metrics in Table~\ref{tab:perm_stress}). Left: reverse-order Turn All-Gold EM shift ($\Delta$\,pp) over answerable queries; reversing input order unexpectedly raises causal-only EM by $+2.76$\,pp on \textsc{LME-S} and $+2.34$\,pp on \textsc{LME-M} because top-ranked turns moved to the end can finally attend to later distractors, whereas \method{} keeps all standard-dialogue shifts within $\pm 0.22$\,pp. Right: reverse-order packed-set Jaccard overlap and Spearman rank correlation $\rho$ across all 4,703 queries; adding the two-layer bidirectional set head improves both packet consistency and rank stability on every benchmark.}
    \label{fig:app_order}
\end{figure*}

\paragraph{Causal Front-Position Blindspot.}
Because \method{} runs its single prefill over a causal language model, candidates placed near the beginning of the prompt cannot attend to candidates appearing later during the backbone layers. If the scoring head reads those causal states directly, top-ranked evidence placed first suffers from a front-position blindspot: it cannot compare itself against later competitors or state updates. To test whether our two-layer bidirectional set head removes this vulnerability, Table~\ref{tab:perm_stress} and Figure~\ref{fig:app_order} evaluate deployed $\tau=0.05$ packing across all 4,703 queries under five orderings: original retrieval order, fully reversed order, and three random shuffles. Alongside Turn All-Gold EM shifts, Table~\ref{tab:perm_stress} tracks reverse-order Spearman $\rho$ and Kendall $\tau_K$ rank correlations, packed-set Jaccard overlap measuring exact identity of admitted turns, gold-turn mean absolute score drift $\text{gMAE}$, and abstention-query token length under normal versus reversed order.

Figure~\ref{fig:app_order} exposes the blindspot sharply. Removing bidirectional interaction in \textsc{CausalMLP-1.7B} causes reversing input order to \emph{raise} Turn All-Gold EM by $+2.76$\,pp on \textsc{LME-S} and $+2.34$\,pp on \textsc{LME-M}, simply because moving top-ranked turns to the back lets them attend to the rest of the pool; meanwhile, abstention packets inflate from 177 to 210 tokens on \textsc{LME-S}. Adding the bidirectional head in \method{} raises original-order EM to $92.55\%$ on \textsc{LME-S} and $85.96\%$ on \textsc{LME-M}, shrinks reverse-order shifts tenfold to $+0.22$ and $+0.21$\,pp, and boosts both Spearman $\rho$ and packed-set Jaccard overlap on every benchmark. Residual order sensitivity remains largest on dense \textsc{UtilMem} at $-1.74$\,pp, where 94.1\% budget binding magnifies minor score fluctuations from the lower causal layers.

\begin{table*}[t]
\centering
\resizebox{\textwidth}{!}{%
\begin{tabular}{@{}ll|ccc|cccccc@{}}
\toprule
\textbf{Benchmark} & \textbf{Model ($\tau{=}0.05$)} & \textbf{Orig.\ EM} & \textbf{Rev.\ EM ($\Delta$)} & \textbf{3-Rand EM ($\Delta$)} & \textbf{Rev.\ $\rho$} & \textbf{Rev.\ $\tau_K$} & \textbf{Rev.\ Jacc.} & \textbf{Rand.\ Jacc.} & \textbf{gMAE} & \textbf{Abs.\ Tok (N/R)} \\
\midrule
\multirow{2}{*}{\textsc{LME-S}}
 & \textsc{CausalMLP-1.7B} & 86.60 & 89.36 ($+2.76$) & 88.16 ($+1.56$) & 0.9016 & 0.7725 & 75.37\% & 81.08\% & 0.1128 & 177 / 210 \\
 & \textbf{\method{} (Ours)} & \textbf{92.55} & \textbf{92.77} ($+0.22$) & \textbf{92.06} ($-0.49$) & \textbf{0.9177} & \textbf{0.8002} & \textbf{78.81\%} & \textbf{83.15\%} & \textbf{0.0935} & 331 / 338 \\
\midrule
\multirow{2}{*}{\textsc{LME-M}}
 & \textsc{CausalMLP-1.7B} & 79.79 & 82.13 ($+2.34$) & 81.35 ($+1.56$) & 0.9125 & 0.7873 & 76.26\% & 80.35\% & 0.0999 & 213 / 225 \\
 & \textbf{\method{} (Ours)} & \textbf{85.96} & \textbf{86.17} ($+0.21$) & \textbf{85.46} ($-0.50$) & \textbf{0.9288} & \textbf{0.8117} & \textbf{78.73\%} & \textbf{82.81\%} & \textbf{0.0847} & 426 / 418 \\
\midrule
\multirow{2}{*}{\textsc{LoCoMo}}
 & \textsc{CausalMLP-1.7B} & 72.34 & 72.79 ($+0.45$) & 72.66 ($+0.32$) & 0.8717 & 0.7319 & 71.98\% & 77.22\% & 0.0857 & 203 / 218 \\
 & \textbf{\method{} (Ours)} & \textbf{74.87} & \textbf{74.68} ($-0.19$) & \textbf{74.77} ($-0.10$) & \textbf{0.9262} & \textbf{0.8089} & \textbf{77.06\%} & \textbf{81.50\%} & \textbf{0.0638} & 316 / 319 \\
\midrule
\multirow{2}{*}{\textsc{UtilMem}}
 & \textsc{CausalMLP-1.7B} & 23.12 & 20.68 ($-2.44$) & 20.81 ($-2.31$) & 0.9139 & 0.7772 & 68.42\% & 71.60\% & \textbf{0.0530} & -- \\
 & \textbf{\method{} (Ours)} & \textbf{26.73} & \textbf{24.99} ($-1.74$) & \textbf{25.04} ($-1.69$) & \textbf{0.9333} & \textbf{0.8142} & \textbf{71.48\%} & \textbf{74.13\%} & 0.0713 & -- \\
\bottomrule
\end{tabular}}
\caption{Deployed candidate-order permutation stress test at $\tau=0.05$ and $k_{\min}=2$ across all 4,703 queries under five orderings: original retrieval order, reversed order, and three random shuffles. \textbf{Bold} marks the better model per benchmark block. EM columns are Turn All-Gold EM (\%) with shifts $\Delta$ in pp relative to original order; $\rho$ and $\tau_K$ are reverse-order Spearman and Kendall rank correlations; Jacc.\ is Jaccard overlap of admitted packs; gMAE is mean absolute admission-score drift on gold turns under reversal; Abs.\ Tok (N/R) is mean evidence tokens on abstention queries under normal versus reversed order, with dashes on \textsc{UtilMem} which contains no abstention queries. Minor Orig.\ EM differences relative to Table~\ref{tab:calib_full} (3/1,536 flips on \textsc{LoCoMo}; 1/1,717 on \textsc{UtilMem}) stem from live unbatched \texttt{bs=1} BF16 re-inference versus batched cached logits.}
\label{tab:perm_stress}
\end{table*}

\paragraph{Unified Budget Sweep and Dense Stratification.}
Two practical deployment questions follow from our packing rule: whether the budget ceiling $B$ requires per-benchmark tuning, and whether exact length-aware dynamic programming outperforms simpler greedy top-probability packing when budgets tighten. Table~\ref{tab:unified_budget} sweeps $B \in \{500, 650, 1000, 1500\}$ at fixed $\tau=0.05$ and confirms that a single universal cap $B=1000$ generalizes cleanly. Raising \textsc{LoCoMo} from $B=650$ to $B=1000$ leaves Turn All-Gold EM unchanged at 75.07\% while adding only 14 tokens because calibrated admission scores stop admitting noise turns well below the ceiling. By contrast, dense \textsc{UtilMem} saturates 94.1\% of queries at $B=1000$ and continues climbing to 36.81\% EM at $B=1500$. Table~\ref{tab:utilmem_strat} further stratifies all 1,717 \textsc{UtilMem} queries by required gold-session count and compares packing solvers. \method{} opens its widest Turn All-Gold EM margins over \textsc{Query-Platt} and \textsc{CausalMLP-1.7B} on queries spanning three or more gold sessions, while exact knapsack dynamic programming beats length-agnostic greedy selection by probability alone by $+3.08$\,pp because it fits more high-value compact turns into a saturated budget.

\paragraph{Explicit Indefinite Pairwise QUBO Probe.}
Our deployed knapsack objective in Eq.~\eqref{eq:knapsack} sums individual turn utilities $u_i = p_i - \tau$ once bidirectional scoring finishes. A natural theoretical concern is whether this additive surrogate misses explicit pairwise synergy between complementary turns or explicit repulsion between redundant duplicates. To test this directly, we construct an indefinite quadratic unconstrained binary optimization (QUBO) probe over frozen post-bidirectional representations $h_i^{(2)} \in \mathbb{R}^{2048}$ and scores $p_i$. Specifically, we project each candidate to $v_i \in \mathbb{R}^{192}$ and train a strictly symmetric pairwise MLP over $[v_i+v_j,\, |v_i-v_j|,\, v_i \odot v_j,\, p_i+p_j,\, |p_i-p_j|,\, p_i p_j,\, e_{ij}]$, where $e_{ij}$ holds \textsc{BGE-M3} cosine similarity, a same-session indicator, and normalized session gap. Passing the symmetrized scalar output through $\tanh$ yields bounded signed affinities $W_{ij} \in (-1, 1)$ without positive-semidefinite constraints: positive $W_{ij}$ rewards complementary co-gold turns, whereas negative $W_{ij}$ penalizes same-session duplicates. Initializing from exact knapsack DP at $\tau=0$, vectorized 1-swap/add/drop local search maximizes:
\begin{equation}
\begin{aligned}
    E(S) &= \sum_{i \in S} p_i + \beta \sum_{\substack{i,j \in S \\ i < j}} \sqrt{p_i p_j}\, W_{ij} \\[-2pt]
         &\quad \text{s.t.}\; \sum_{i \in S} \ell_i \le B.
\end{aligned}
\label{eq:qubo}
\end{equation}
As Table~\ref{tab:pairwise_probe} shows, the frozen probe separates complementary co-gold pairs from same-session redundancies at 97.3\%--98.8\% zero-shot AUROC on standard dialogue and 89.7\% on OOD \textsc{UtilMem}, confirming that pairwise structure is already encoded inside $h_i^{(2)}$. Nevertheless, synthetic validation selects $\beta^* = 0.0$: activating explicit pairwise weights $\beta > 0$ slightly lowers standard-dialogue EM while adding only $+0.41$\,pp on \textsc{UtilMem}. This result empirically justifies keeping Eq.~\eqref{eq:knapsack} additive and globally optimal via 0.2\,ms exact DP rather than approximate quadratic search.

\begin{table*}[t]
\centering
\resizebox{0.92\textwidth}{!}{%
\begin{tabular}{@{}lcccc@{}}
\toprule
\textbf{Benchmark} & \textbf{$B{=}500$} & \textbf{$B{=}650$} & \textbf{$B{=}1000$} & \textbf{$B{=}1500$} \\
\midrule
\textsc{LoCoMo} ($n{=}1{,}536$) & 74.80 / 291t (18.0\%) & 75.07 / 312t (8.5\%) & \textbf{75.07} / 326t (1.0\%) & 75.07 / 328t (0.0\%) \\
\textsc{LME-S} ($n{=}470$)  & 91.06 / 333t (26.0\%) & -- & \textbf{92.55} / 406t (7.7\%) & -- \\
\textsc{LME-M} ($n{=}470$)  & 84.04 / 369t (36.6\%) & -- & \textbf{85.96} / 467t (9.4\%) & -- \\
\textsc{UtilMem} ($n{=}1{,}717$) & 13.92 / 483t (99.7\%) & -- & 26.67 / 964t (94.1\%) & \textbf{36.81} / 1,400t (82.4\%) \\
\bottomrule
\end{tabular}}
\caption{Unified token-budget sweep for \method{} at $\tau=0.05$ and $k_{\min}=2$. Each cell reports Turn All-Gold EM (\%) / mean selected evidence tokens (budget binding rate \%). \textbf{Bold} marks the highest EM per row, and dashes denote unprofiled budget points.}
\label{tab:unified_budget}

\vspace{7pt}
\resizebox{0.82\textwidth}{!}{%
\begin{tabular}{@{}lccc@{}}
\toprule
\textbf{\textsc{UtilMem} Regime ($B{=}1000$, $\tau{=}0.05$)} & \textbf{\textsc{Query-Platt}} & \textbf{\textsc{CausalMLP-1.7B}} & \textbf{\method{} (Ours)} \\
\midrule
3--4 gold sessions & 27.53\% & 38.67\% & \textbf{43.53\%} \\
5+ gold sessions   & 4.81\%  & 10.15\% & \textbf{12.61\%} \\
\midrule
\textbf{Packing Solver Comparison (All $n{=}1{,}717$)} & \textbf{Greedy by $p_i$} & \textbf{Exact Knapsack DP (Eq.~\ref{eq:knapsack})} & \textbf{Gain ($\Delta$)} \\
\midrule
\method{} scores ($B{=}1000$, $\tau{=}0.05$) & 23.59\% & \textbf{26.67\%} & \textbf{+3.08\,pp} \\
\bottomrule
\end{tabular}}
\caption{\textsc{UtilMem} high-density gold-session stratification (top) and exact knapsack dynamic programming versus length-agnostic greedy selection by probability alone on all 1,717 queries (bottom). \textbf{Bold} marks the highest Turn All-Gold EM per row.}
\label{tab:utilmem_strat}

\vspace{7pt}
\resizebox{\textwidth}{!}{%
\begin{tabular}{@{}lcccccc@{}}
\toprule
\textbf{Benchmark} & \textbf{Co-Gold Mean $W$} & \textbf{Redundant Mean $W$} & \textbf{Co-Gold vs.\ Red.\ AUROC} & \textbf{EM ($\beta^* {=} 0.0$)} & \textbf{EM ($\beta {=} 0.05$)} & \textbf{EM ($\beta {=} 0.10$)} \\
\midrule
Synthetic Val ($n{=}687$) & $+0.795$ & $-0.724$ & 99.13\% & \textbf{89.52\%} & 89.52\% & 89.52\% \\
\textsc{LoCoMo} ($B{=}650$) & $+0.642$ & $-0.712$ & 97.29\% & \textbf{75.91\%} & 75.91\% & 75.85\% \\
\textsc{LME-S} ($B{=}1000$) & $+0.774$ & $-0.700$ & 98.77\% & \textbf{95.74\%} & 95.53\% & 95.11\% \\
\textsc{LME-M} ($B{=}1000$) & $+0.773$ & $-0.683$ & 98.57\% & \textbf{88.30\%} & 88.09\% & 87.87\% \\
\textsc{UtilMem} ($B{=}1000$)& $+0.429$ & $-0.448$ & 89.70\% & 34.94\% & 35.24\% & \textbf{35.35\%} \\
\bottomrule
\end{tabular}}
\caption{Explicit indefinite (non-PSD) symmetric pairwise QUBO probe (Eq.~\ref{eq:qubo}) over frozen post-bidirectional representations at $\tau=0$. $W \in (-1, 1)$ is the mean learned affinity for co-gold versus same-session redundant pairs; AUROC measures pairwise separation; EM columns report Turn All-Gold EM (\%) across pairwise weights $\beta$, with \textbf{bold} marking the highest EM per row. Validation selects $\beta^*=0.0$.}
\label{tab:pairwise_probe}
\end{table*}

\section{Complete Five-Reader Matrix, Statistical Protocol, and Fairness Audits}
\label{app:qa_ablations}

\subsection{Statistical Testing Protocol}

All significance tests compare paired predictions over identical question subsets so gains cannot arise from sample mismatch. For single-reader binary QA comparisons, we apply the two-sided exact McNemar test over discordant pairs, which tests whether one system flips strictly more questions from wrong to right than the reverse. For multi-reader pooled accuracy on \textsc{LME-S} and \textsc{LME-M} and continuous 0--100 rubric scores on \textsc{UtilMem}, we compute two-sided $p$-values and 95\% confidence intervals via 10,000-resample clustered bootstrap \citep{efron1993bootstrap}. Resampling whole questions as clusters preserves within-question correlation across readers and avoids overstating effective sample size. On \textsc{LoCoMo}, where 1,540 answerable queries share ten underlying dialogues, we resample whole dialogues via a 10-conversation cluster bootstrap and additionally report leave-one-conversation-out ranges to verify that no single conversation drives the margin. Within each multi-baseline comparison family, raw $p$-values are adjusted using the Holm--Bonferroni step-down procedure \citep{holm1979simple} to control family-wise false positives.

\subsection{Full Reader Matrices, Selector Suite, and Category Breakdowns}

\begin{figure*}[t]
    \centering
    \includegraphics[width=0.98\textwidth]{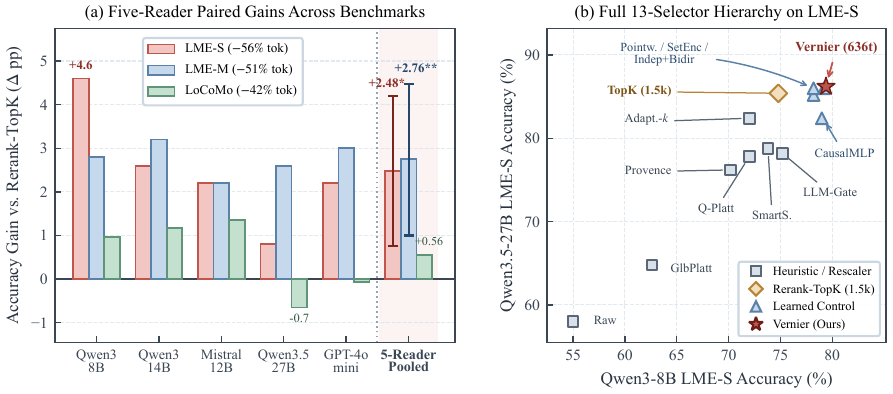}
    \caption{Cross-reader generalization and full 13-selector hierarchy (complete tables in Tables~\ref{tab:full_5reader}--\ref{tab:platt_qa}). Left: paired accuracy gain ($\Delta$\,pp) of \method{} over full-budget \textsc{Rerank-TopK} across all five readers and the 5-reader pooled summary on \textsc{LME-S}, \textsc{LME-M}, and \textsc{LoCoMo} Cat 1--4; error bars denote 95\% clustered bootstrap CIs on pooled \textsc{LME-S} ($+2.48$\,pp, $^*$\,Holm $p=.0136$) and \textsc{LME-M} ($+2.76$\,pp, $^{**}$\,Holm $p=.0064$). Right: \textsc{Qwen3-8B} versus \textsc{Qwen3.5-27B} accuracy across all 13 query-time selectors on \textsc{LME-S} ($n=500$); all heuristic cutoffs and post-hoc rescalers cluster well below learned neural selectors, and \method{} reaches the Pareto-dominant upper-right corner at 636 mean reader tokens.}
    \label{fig:app_readers}
\end{figure*}

\paragraph{Five-Reader Query-Time Matrix.}
To verify that evidence pruning helps across model scales, model families, and serving regimes rather than overfitting to one backbone, Figure~\ref{fig:app_readers}(a) and Table~\ref{tab:full_5reader} report all 15 reader--benchmark combinations under unified \textsc{DeepSeek-V4.1-Flash} judging. Alongside our primary open readers, we include \textsc{Qwen3-14B} as an intermediate parameter-scaling check, \textsc{Mistral-12B} as a cross-family open reader, and \textsc{GPT-4o-mini} as a closed API reader. Because \textsc{Qwen3-14B} scored slightly below \textsc{Qwen3-8B} across every selector on \textsc{LongMemEval}, we evaluated it on the four core zero-build arms; \method{} improves \textsc{Qwen3-14B} on all three benchmarks by $+2.60$\,pp on \textsc{LME-S}, $+3.20$\,pp on \textsc{LME-M}, and $+1.17$\,pp on \textsc{LoCoMo}. Across all five readers, pooled 500-question clustered bootstrap confirms that \method{} beats \textsc{Rerank-TopK} by $+2.48$\,pp on \textsc{LME-S}, 95\% CI $[+0.76, +4.20]$ (Holm $p=0.0136$), and by $+2.76$\,pp on \textsc{LME-M}, 95\% CI $[+1.00, +4.48]$ (Holm $p=0.0064$), while cutting reader tokens by 51\%--56\%.

\paragraph{Full 13-Selector Suite and Dual-Judge Cross-Check.}
To separate genuine answer-oriented set selection from cheaper threshold heuristics, post-hoc rescalers, and alternative neural designs, Figure~\ref{fig:app_readers}(b) and Tables~\ref{tab:ans_abs}--\ref{tab:platt_qa} compare all 13 query-time selectors sharing our frozen Top-32 pool on \textsc{LongMemEval-S}. Table~\ref{tab:ans_abs} disaggregates overall accuracy into the 470 answerable and 30 abstention queries alongside 27B Multi-Session and Temporal-Reasoning subsets. Two clear patterns emerge. First, every heuristic cutoff and post-hoc calibrator trails learned selectors across both 8B and 27B readers, losing up to 14--28\,pp on Multi-Session and Temporal-Reasoning queries where multiple dated turns must survive together. Second, among learned neural selectors, \method{} leads on both readers simultaneously while reaching the highest 27B Multi-Session score of 81.0\%. On 1.5M-token \textsc{LME-M} with \textsc{Qwen3.5-27B}, \method{} further leads \textsc{Set-Encoder}-style by $+3.6$\,pp, reaching 79.4\% versus 75.8\% (raw McNemar $p=0.013$, 4-comparison Holm $p=0.053$). Table~\ref{tab:platt_qa} additionally verifies that all 18 \textsc{LME-S} control rankings hold identically under both external \textsc{DeepSeek} and local deterministic \textsc{Qwen3.5-27B} judges, and includes a token-matched \textsc{CausalMLP-1.7B} control at $\tau=0.025$ to separate cutoff tightness from bidirectional interaction.

\paragraph{\textsc{LoCoMo-1986} Categories and Prompt Fairness.}
Because \textsc{LoCoMo-1986} mixes five distinct reasoning categories including adversarial unanswerable traps, Table~\ref{tab:locomo_cat} disaggregates performance across all five categories and all five readers. \method{} improves Multi-Hop and Temporal accuracy across most readers at 42\% fewer tokens and raises 8B Category-5 abstention to 57.40\% versus 51.57\% for \textsc{Rerank-TopK} and 48.88\% at $\tau=0$, confirming that pruning false-premise noise reduces hallucination on smaller readers. Across all ten dialogues, leave-one-conversation-out deltas remain strictly positive against \textsc{MemLoc-4B} on 10/10 folds, ranging from $+15.63$ to $+17.67$\,pp on 8B and $+16.38$ to $+17.34$\,pp on 27B; against \textsc{FacLoc} on 10/10 folds, from $+6.84$ to $+8.15$\,pp on 8B and $+8.05$ to $+8.89$\,pp on 27B; and against 8B \textsc{FullContext} on 9/10 folds, from $+3.31$ to $+5.26$\,pp. For strict baseline fairness, evaluating \textsc{MemLoc-4B} under our symmetric reader prompt lowered its Category 1--4 accuracy to 52.40\% on 8B and 45.97\% on 27B, below the official prompt scores of 58.77\% and 59.94\% reported in Table~\ref{tab:pareto} and \S\ref{sec:rq1}; we therefore report its stronger official prompt. Finally, we omit a \textsc{LoCoMo} Gold-Only Oracle because gold \texttt{dia\_id} annotations isolate bare single utterances averaging 1.50 sentences and 72.6 tokens without surrounding antecedent turns, dropping isolated-gold accuracy to 68.25\%.

\begin{table*}[t]
\centering
\resizebox{\textwidth}{!}{%
\begin{tabular}{@{}llcccccccc@{}}
\toprule
\textbf{Reader} & \textbf{Bench.} & \textbf{NaiveRAG} & \textbf{FacLoc} & \textbf{Raw ($\tau{=}.05$)} & \textbf{TopK (pTok)} & \textbf{CausalMLP} & \textbf{Vernier ($\tau{=}0$)} & \textbf{Vernier (pTok)} & \textbf{w/o Date} \\
\midrule
\textsc{Qwen3-8B}       & LME-S & 67.60 & 72.00 & 55.00 & 74.80 (1,449) & 79.00 & 76.60 & \textbf{79.40} (\textbf{636}) & 78.00 \\
\textsc{Qwen3-14B}      & LME-S & 68.40 & 71.40 & --    & 74.60 (1,449) & --    & --    & \textbf{77.20} (\textbf{636}) & --    \\
\textsc{Mistral-12B}    & LME-S & 68.40 & 69.40 & 54.80 & 74.20 (1,477) & 75.40 & 75.40 & \textbf{76.40} (\textbf{642}) & 77.00 \\
\textsc{Qwen3.5-27B}    & LME-S & 77.00 & 83.00 & 58.00 & 85.40 (1,478) & 82.40 & 85.60 & \textbf{86.20} (\textbf{648}) & 85.00 \\
\textsc{GPT-4o-mini}    & LME-S & 75.20 & 80.20 & 57.80 & 83.60 (1,310) & 82.60 & 82.40 & \textbf{85.80} (\textbf{579}) & 85.20 \\
\midrule
\textsc{Qwen3-8B}       & LME-M & 60.40 & 60.40 & 51.00 & 68.80 (1,563) & \textbf{73.00} & 70.00 & 71.60 (\textbf{769}) & 70.60 \\
\textsc{Qwen3-14B}      & LME-M & 60.40 & 59.60 & --    & 67.60 (1,563) & --    & --    & \textbf{70.80} (\textbf{769}) & --    \\
\textsc{Mistral-12B}    & LME-M & 63.40 & 62.40 & 51.80 & 67.20 (1,593) & 68.00 & 66.60 & \textbf{69.40} (\textbf{777}) & 69.40 \\
\textsc{Qwen3.5-27B}    & LME-M & 70.80 & 73.20 & 55.20 & 76.80 (1,592) & 78.00 & 78.60 & \textbf{79.40} (\textbf{783}) & 76.60 \\
\textsc{GPT-4o-mini}    & LME-M & 68.00 & 69.80 & 55.40 & 75.00 (1,399) & 77.00 & 73.20 & \textbf{78.00} (\textbf{694}) & 78.00 \\
\midrule
\textsc{Qwen3-8B}       & LoCoMo & 65.32 & 68.05 & 72.47 & 74.61 (1,262) & 75.39 & 75.06 & \textbf{75.58} (\textbf{725}) & 72.99 \\
\textsc{Qwen3-14B}      & LoCoMo & 65.00 & 67.21 & --    & 75.84 (1,262) & --    & --    & \textbf{77.01} (\textbf{725}) & --    \\
\textsc{Mistral-12B}    & LoCoMo & 58.57 & 61.04 & 65.19 & 69.68 (1,267) & 70.19 & 69.81 & \textbf{71.04} (\textbf{722}) & 67.79 \\
\textsc{Qwen3.5-27B}    & LoCoMo & 65.78 & 68.44 & 72.66 & 77.53 (1,295) & 75.71 & \textbf{78.70} & 76.88 (\textbf{744}) & 74.42 \\
\textsc{GPT-4o-mini}    & LoCoMo & 62.73 & 66.23 & 71.04 & \textbf{74.94} (1,136) & 74.81 & 74.42 & 74.87 (\textbf{661}) & 72.21 \\
\bottomrule
\end{tabular}}
\caption{Complete 5-reader $\times$ 8-arm query-time QA accuracy (\%) under unified \textsc{DeepSeek-V4.1-Flash} judging on \textsc{LME-S} ($n=500$), \textsc{LME-M} ($n=500$), and \textsc{LoCoMo} Cat 1--4 ($n=1{,}540$). \textbf{Bold} marks the highest accuracy per row and the lower mean reader prompt token count (pTok) between \textsc{Rerank-TopK} and \method{}. Columns show raw reranker thresholding (\textsc{Raw}), \textsc{Rerank-TopK} (\textsc{TopK}), \textsc{CausalMLP-1.7B}, unthresholded \method{} ($\tau=0$), default \method{} ($\tau=0.05$), and \method{} without session-date headers (\textsc{w/o Date}). Dashes indicate ablation arms omitted on the \textsc{Qwen3-14B} scaling control.}
\label{tab:full_5reader}

\vspace{7pt}
\resizebox{\textwidth}{!}{%
\begin{tabular}{@{}l|cccc|cccccc@{}}
\toprule
\multirow{2}{*}{\textbf{Selector (\textsc{LongMemEval-S}, \textsc{DeepSeek} Judge)}} & \multicolumn{4}{c|}{\textbf{\textsc{Qwen3-8B} Reader}} & \multicolumn{6}{c}{\textbf{\textsc{Qwen3.5-27B} Reader}} \\
 & \textbf{All (500)} & \textbf{Ans (470)} & \textbf{Abs (30)} & \textbf{pTok} & \textbf{All (500)} & \textbf{Ans (470)} & \textbf{Abs (30)} & \textbf{MS (121)} & \textbf{TR (127)} & \textbf{pTok} \\
\midrule
\textsc{Rerank-TopK} ($B{=}1000$)                  & 74.8 & 73.6 & 93.3 & 1,449 & 85.4 & 84.9 & 93.3 & 79.3 & 80.3 & 1,478 \\
\textsc{Raw Reranker} ($\tau{=}0.05$)              & 55.0 & 52.3 & \textbf{96.7} & \textbf{386} & 58.0 & 55.7 & 93.3 & 25.6 & 32.3 & \textbf{393} \\
\textsc{Global Platt} ($\tau{=}0.05$)              & 62.6 & 60.4 & \textbf{96.7} & 457 & 64.8 & 62.8 & \textbf{96.7} & 41.3 & 41.7 & 466 \\
\textsc{Query-Platt} ($\tau{=}0.05$)               & 72.0 & 70.4 & \textbf{96.7} & 567 & 77.8 & 76.6 & \textbf{96.7} & 60.3 & 69.3 & 577 \\
\textsc{Adaptive-$k$} (RRF, $B{=}5$)               & 72.0 & 70.6 & 93.3 & 953 & 82.4 & 81.5 & \textbf{96.7} & 71.1 & 79.5 & 973 \\
\textsc{Provence} ($T{=}0.1$)                      & 70.2 & 68.9 & 90.0 & 527 & 76.2 & 75.5 & 86.7 & 65.3 & 63.8 & 536 \\
\textsc{SmartSearch}-style ($\alpha{=}0.02$)       & 73.8 & 72.3 & \textbf{96.7} & 679 & 78.8 & 77.7 & \textbf{96.7} & 62.0 & 71.7 & 692 \\
\textsc{LLM Recall Gating} (\textsc{Qwen3-8B})     & 75.2 & 74.3 & 90.0 & 481 & 78.2 & 77.4 & 90.0 & 63.6 & 74.8 & 490 \\
\midrule
\textsc{Set-Encoder}-style (435M)                  & 78.2 & 77.2 & 93.3 & 595 & \ul{86.0} & \ul{85.5} & 93.3 & 78.5 & 84.3 & 605 \\
\textsc{Pointwise-1.7B} ($\tau{=}0.05$)            & 78.2 & 77.4 & 90.0 & 611 & 85.2 & 84.9 & 90.0 & 73.6 & 85.0 & 622 \\
\textsc{CausalMLP-1.7B} ($\tau{=}0.05$)            & 79.0 & \textbf{78.5} & 86.7 & 491 & 82.4 & 81.9 & 90.0 & 72.7 & 81.1 & 500 \\
\textsc{Indep+Bidir} (1.7B, $\tau{=}0.05$)         & \ul{79.2} & \textbf{78.5} & 90.0 & 616 & \ul{86.0} & 85.3 & \textbf{96.7} & 76.9 & \textbf{85.8} & 627 \\
\textbf{\method{} (Ours, $\tau{=}0.05$)}           & \textbf{79.4} & \textbf{78.5} & 93.3 & 636 & \textbf{86.2} & \textbf{85.7} & 93.3 & \textbf{81.0} & \ul{85.0} & 648 \\
\bottomrule
\end{tabular}}
\caption{Answerable versus abstention accuracy (\%) on \textsc{LongMemEval-S} across all 13 query-time selectors under unified \textsc{DeepSeek} judging. \textbf{Bold} and \ul{underline} mark best and second-best values per column; pTok is mean reader prompt tokens; MS and TR report Multi-Session and Temporal-Reasoning accuracy on 27B. Because only 30 abstention queries exist, a single query corresponds to 3.3\,pp. Under the unified 18-arm Holm family, \method{} leads \textsc{Provence} ($T=0.1$) at $p<.0001$ on both readers and \textsc{Adaptive-$k$} at $p=.0015$ on 8B ($p=.1144$ on 27B).}
\label{tab:ans_abs}
\end{table*}

\begin{table*}[t]
\centering
\resizebox{\textwidth}{!}{%
\begin{tabular}{@{}l|ccc|ccc@{}}
\toprule
\multirow{2}{*}{\textbf{Selector (\textsc{LongMemEval-S}, $n{=}500$)}} & \multicolumn{3}{c|}{\textbf{\textsc{Qwen3-8B} Reader}} & \multicolumn{3}{c}{\textbf{\textsc{Qwen3.5-27B} Reader}} \\
 & \textbf{pTok} & \textbf{DS Acc.\ ($\Delta$, Holm $p$)} & \textbf{Loc-27B Acc.} & \textbf{pTok} & \textbf{DS Acc.\ ($\Delta$, Holm $p$)} & \textbf{Loc-27B Acc.} \\
\midrule
\textsc{BaseProv Raw} ($\tau{=}0.05$) & 386.0 & 55.00 ($+24.40$, $<.0001$) & 53.40 & 393.0 & 58.00 ($+28.20$, $<.0001$) & 56.60 \\
\textsc{Global Platt} ($\tau{=}0.05$) & 457.0 & 62.60 ($+16.80$, $<.0001$) & 61.20 & 465.5 & 64.80 ($+21.40$, $<.0001$) & 64.60 \\
\textsc{Query-Platt} ($\tau{=}0.05$) & 566.5 & 72.00 ($+7.40$, $.0007$) & 69.60 & 577.2 & 77.80 ($+8.40$, $<.0001$) & 76.20 \\
\textsc{SmartSearch}-style ($\alpha{=}0.02$, val-matched) & 678.7 & 73.80 ($+5.60$, $.0417$) & 71.40 & 691.8 & 78.80 ($+7.40$, $<.0001$) & 77.40 \\
\textsc{SmartSearch}-style ($\alpha{=}0.01$, +18\% tok) & 752.3 & 75.00 ($+4.40$, $.1784$) & 71.60 & 767.1 & 80.40 ($+5.80$, $.0024$) & 79.00 \\
\textsc{LLM Recall Gating} (\textsc{Qwen3-1.7B}, 198\,ms) & 889.1 & 66.00 ($+13.40$, $<.0001$) & 62.40 & 905.8 & 69.20 ($+17.00$, $<.0001$) & 67.20 \\
\textsc{LLM Recall Gating} (\textsc{Qwen3-8B}, 226\,ms) & 481.2 & 75.20 ($+4.20$, $.2836$) & 71.00 & 489.6 & 78.20 ($+8.00$, $<.0001$) & 75.20 \\
\textsc{Rerank-TopK} (static $B{=}500$ / full $B{=}1000$) & 786 / 1,449 & 73.00 / 74.80 & 69.80 / 70.60 & -- / 1,478 & -- / 85.40 & -- / 82.80 \\
\textsc{Pointwise-1.7B} ($\tau{=}0.05$, 65\,ms KV) & 611.0 & 78.20 ($+1.20$, $1.0$) & 74.40 & 622.0 & 85.20 ($+1.00$, $1.0$) & 81.80 \\
\midrule
\textsc{CausalMLP-1.7B} ($\tau{=}0.05$, default) & 491.4 & 79.00 ($+0.40$, $1.0$) & 75.60 & 499.8 & 82.40 ($+3.80$, $.0635$) & 79.40 \\
\textsc{CausalMLP-1.7B} ($\tau{=}0.025$, tok-matched) & 652.8 & 78.60 ($+0.80$, $1.0$) & 75.80 & 664.6 & \textbf{87.00} ($-0.80$, $1.0$) & \textbf{84.60} \\
\midrule
\textbf{\method{} ($\tau{=}0.08$, compact)} & \textbf{527.6} & \textbf{79.40} ($0.00$, $1.0$) & \textbf{76.40} & \textbf{536.7} & 84.60 ($+1.60$, $1.0$) & 80.80 \\
\textbf{\method{} ($\tau{=}0.05$, default, 54\,ms)} & \textbf{636.1} & \textbf{79.40} (ref.) & 76.00 & \textbf{647.8} & 86.20 (ref.) & 82.80 \\
\midrule
\emph{Gold-Only Oracle Reference (Top-32)} & \emph{372.2} & \emph{78.40} & \emph{76.80} & -- & -- & -- \\
\bottomrule
\end{tabular}}
\caption{Complete end-to-end QA accuracy (\%) controls on \textsc{LongMemEval-S} ($n=500$) under external \textsc{DeepSeek} (DS) and local deterministic \textsc{Qwen3.5-27B} (Loc-27B) judges. \textbf{Bold} highlights the highest accuracy per column and \method{}'s prompt token counts (pTok). $\Delta$ (pp) and unified 18-arm Holm-adjusted exact McNemar $p$-values across all \textsc{LME-S} controls compare \method{} ($\tau=0.05$, ref.) against each selector; KV denotes prefix KV-caching, and tok-matched tunes \textsc{CausalMLP-1.7B} to match \method{}'s prompt budget. Gold-Only Oracle retains only gold turns present in Top-32; dashes mark arms evaluated solely on 8B.}
\label{tab:platt_qa}

\vspace{7pt}
\resizebox{\textwidth}{!}{%
\begin{tabular}{@{}llccccccc@{}}
\toprule
\textbf{Reader} & \textbf{Method} & \textbf{Single-Hop (841)} & \textbf{Multi-Hop (282)} & \textbf{Temporal (321)} & \textbf{Open-Dom.\ (96)} & \textbf{Cat-5 Adv.\ (446)} & \textbf{All-1986} & \textbf{pTok} \\
\midrule
\multirow{3}{*}{\textsc{Qwen3-8B}}
 & \textsc{BaseProv FacLoc} & 79.19 & 64.89 & 50.16 & 39.58 & 50.90 & 64.20 & 1,344 \\
 & \textsc{Rerank-TopK}     & \textbf{87.63} & 68.44 & 57.01 & 37.50 & 51.57 & 69.44 & 1,262 \\
 & \textbf{\method{} ($\tau{=}0.05$)} & 86.68 & \textbf{70.92} & \textbf{61.06} & \textbf{40.62} & \textbf{57.40} & \textbf{71.50} & \textbf{725} \\
\midrule
\multirow{2}{*}{\textsc{Qwen3-14B}}
 & \textsc{Rerank-TopK}     & 86.09 & 68.44 & 66.36 & 39.58 & \textbf{66.82} & \textbf{73.82} & 1,262 \\
 & \textbf{\method{} ($\tau{=}0.05$)} & \textbf{86.56} & \textbf{72.34} & \textbf{66.67} & \textbf{41.67} & 61.66 & 73.56 & \textbf{725} \\
\midrule
\multirow{2}{*}{\textsc{Mistral-12B}}
 & \textsc{Rerank-TopK}     & \textbf{83.00} & 60.99 & 54.83 & 28.12 & \textbf{79.60} & \textbf{71.90} & 1,267 \\
 & \textbf{\method{} ($\tau{=}0.05$)} & 82.52 & \textbf{64.18} & \textbf{58.88} & \textbf{31.25} & 73.09 & 71.50 & \textbf{722} \\
\midrule
\multirow{3}{*}{\textsc{Qwen3.5-27B}}
 & \textsc{Rerank-TopK}     & 87.04 & 67.73 & 73.21 & 37.50 & \textbf{93.72} & 81.17 & 1,295 \\
 & \textsc{Vernier ($\tau{=}0$)} & \textbf{87.75} & \textbf{71.28} & \textbf{74.45} & 35.42 & 93.05 & \textbf{81.92} & 1,338 \\
 & \textbf{\method{} ($\tau{=}0.05$)} & 85.73 & 68.09 & 72.90 & \textbf{38.54} & 91.93 & 80.26 & \textbf{744} \\
\midrule
\multirow{2}{*}{\textsc{GPT-4o-mini}}
 & \textsc{Rerank-TopK}     & \textbf{85.97} & 63.12 & \textbf{70.40} & 28.12 & 78.25 & 75.68 & 1,136 \\
 & \textbf{\method{} ($\tau{=}0.05$)} & 84.66 & \textbf{65.60} & 70.09 & \textbf{32.29} & \textbf{78.92} & \textbf{75.78} & \textbf{661} \\
\bottomrule
\end{tabular}}
\caption{Complete \textsc{LoCoMo-1986} five-category and overall accuracy (\%) across all five readers under unified \textsc{DeepSeek} judging, with per-category query counts in headers. \textbf{Bold} marks the best score and lowest mean reader prompt tokens (pTok) per reader block. \textsc{Qwen3-8B} includes \textsc{FacLoc} to contrast similarity repulsion on Cat-5, and \textsc{Qwen3.5-27B} includes $\tau=0$ to isolate the cutoff on the strongest open reader.}
\label{tab:locomo_cat}
\end{table*}

\section{System Profiling, Tri-Judge Validation, and Blind Audit}
\label{app:profiling}

\subsection{Single-GPU Stage-by-Stage Latency Breakdown}

A practical concern with adding a 1.7B neural selector is whether Stage~2 slows down query serving or requires heavy memory. To measure worst-case wall-clock cost, Table~\ref{tab:latency_detail} profiles every offline and online stage on one NVIDIA H800 GPU at batch size 1 without prefix caching, averaged across 50 selection queries, 30 \textsc{Qwen3-8B} generations, and 10 full 1.5M-token \textsc{LME-M} histories averaging 4,922 turns. Offline non-LLM indexing requires only 0.30--8.04\,s and 1.46--18.99\,MB per history, shrinking disk footprint 436$\times$ relative to \textsc{MemLoc-4B} while using zero LLM tokens. Online, \method{} completes its single zero-decode prefill in 45.0--53.9\,ms within 7.25\,GB peak VRAM and runs exact CPU dynamic programming in roughly 0.2\,ms. That single prefill runs 3.7--4.2$\times$ faster than generative \textsc{LLM Recall Gating} and stays faster than independent \textsc{Pointwise-1.7B} scoring even with KV caching. Most importantly, pruning reader prompts saves 58.4\,ms of \textsc{Qwen3-8B} latency on \textsc{LME-S}, fully offsetting Stage~2 and matching \textsc{Rerank-TopK} end-to-end latency at 1.55\,s.

\begin{table*}[t]
\centering
\resizebox{0.92\textwidth}{!}{%
\begin{tabular}{@{}lccc@{}}
\toprule
\textbf{Stage / Resource} & \textbf{\textsc{LoCoMo} (22k)} & \textbf{\textsc{LME-S} (110k)} & \textbf{\textsc{LME-M} (1.5M)} \\
\midrule
Offline BGE-M3 + TF-IDF (0 LLM tok) & 0.30\,s / hist. & 1.36\,s / hist. & 8.04\,s / hist. \\
Offline Index Disk Footprint & 1.46\,MB / hist. & 1.94\,MB / hist. & 18.99\,MB / hist. \\
\midrule
Stage 1a: RRF $\rightarrow$ Pool-80 (p50 / p95) & 12.6 / 17.0\,ms & 13.5 / 22.8\,ms & 45.4\,ms (p50) \\
Stage 1b: \textsc{BaseProv}-80 $\rightarrow$ Top-32 & 111.6 / 134.8\,ms & 341.3 / 493.0\,ms & 493.1\,ms (p50) \\
Stage 2: \method{} 1.7B Prefill & 48.7 / 64.2\,ms & 53.9 / 64.6\,ms & 45.0 / 52.8\,ms \\
Stage 3: Exact Knapsack DP (CPU) & 0.23 / 0.30\,ms & 0.21 / 0.34\,ms & 0.18\,ms (p50) \\
\textbf{Total Online Selection (p50 / p95)} & \textbf{174.8 / 206.7\,ms} & \textbf{411.6 / 573.9\,ms} & \textbf{583.7\,ms} (sum) \\
\midrule
\textit{Alternative Selectors (Stage 2 Only, \texttt{bs=1})} & & & \\
\textsc{Set-Encoder}-style (435M) & 34.2 / 42.6\,ms & 36.8 / 39.5\,ms & -- \\
\textsc{Pointwise-1.7B} (KV-cache / merged uncached) & 67.2 / 58.0\,ms & 65.1 / 71.2\,ms & -- \\
\textsc{Indep+Bidir} (1.7B, 8.86\,GB) & 60.4 / 64.3\,ms & 74.4 / 78.0\,ms & -- \\
\textsc{LLM Gate} (\textsc{Qwen3-1.7B} / 8B) & -- & 198.0 / 226.1\,ms & -- \\
\midrule
\textit{Back-to-Back \textsc{Qwen3-8B} Reader + Total E2E} & & & \\
\textsc{Rerank-TopK} Reader / Total E2E & 117.1\,ms / \textbf{0.24\,s} & 1,198.6\,ms / 1.55\,s & -- / 0.54\,s (sel.) \\
\textbf{\method{} Reader / Total E2E (p50)} & \textbf{108.1\,ms} / 0.28\,s & \textbf{1,140.2\,ms} / \textbf{1.55\,s} & -- / 0.58\,s (sel.) \\
\bottomrule
\end{tabular}}
\caption{Single-GPU 1$\times$H800 \texttt{bs=1} stage-by-stage latency across 22k--1.5M histories without prefix caching. \textbf{Bold} marks faster reader or end-to-end timings in the bottom block and total online selection. Entries show median (p50) and 95th-percentile (p95) latency; (sum) sums stage medians on \textsc{LME-M}, (sel.) denotes selection-only timing, and dashes indicate unprofiled cells. Peak VRAM is 8.91\,GB for Stage 1a+1b and 7.25\,GB for merged \method{}; unmerged \textsc{Pointwise-1.7B} runs in 64.9\,ms on \textsc{LoCoMo} and 82.1\,ms on \textsc{LME-S}.}
\label{tab:latency_detail}

\vspace{7pt}
\resizebox{0.78\textwidth}{!}{%
\begin{tabular}{@{}llccc@{}}
\toprule
\textbf{Benchmark $\times$ Reader} & \textbf{Judge Pair} & \textbf{Paired $N$} & \textbf{Binary Agree (\%)} & \textbf{Reliability ($\kappa$ / $r$)} \\
\midrule
\textsc{LME-S} $\times$ \textsc{Qwen3-8B}       & \textsc{GPT-4o} vs.\ \textsc{DeepSeek}     & 8,000 & 95.97\% & $\kappa{=}0.897$ \\
\textsc{LME-S} $\times$ \textsc{Qwen3-8B}       & Local-27B vs.\ \textsc{GPT-4o}             & 8,000 & 96.75\% & $\kappa{=}0.920$ \\
\textsc{LME-S} $\times$ \textsc{Qwen3-8B}       & Local-27B vs.\ \textsc{DeepSeek}           & 8,000 & 96.12\% & $\kappa{=}0.903$ \\
\textsc{LME-S} $\times$ \textsc{GPT-4o-mini}    & Local-27B vs.\ \textsc{GPT-4o}             & 5,500 & 97.18\% & $\kappa{=}0.916$ \\
\textsc{LME-S} $\times$ \textsc{GPT-4o-mini}    & \textsc{GPT-4o} vs.\ \textsc{DeepSeek}     & 5,500 & 96.56\% & $\kappa{=}0.893$ \\
\textsc{LME-M} $\times$ \textsc{Qwen3-8B}       & \textsc{GPT-4o} vs.\ \textsc{DeepSeek}     & 8,000 & 95.58\% & $\kappa{=}0.900$ \\
\textsc{LME-M} $\times$ \textsc{GPT-4o-mini}    & \textsc{GPT-4o} vs.\ \textsc{DeepSeek}     & 5,500 & 97.36\% & $\kappa{=}0.933$ \\
\textsc{UtilMem} $\times$ \textsc{Qwen3-8B}     & Local-27B vs.\ \textsc{DeepSeek}           & 5,391 & 94.01\% & $r{=}0.826$ \\
\textsc{UtilMem} $\times$ \textsc{GPT-4o-mini}  & Local-27B vs.\ \textsc{DeepSeek}           & 3,048 & 89.90\% & $r{=}0.832$ \\
\bottomrule
\end{tabular}}
\caption{Inter-judge reliability across \textsc{GPT-4o-2024-11-20}, \textsc{DeepSeek-V4.1-Flash}, and deterministic local \textsc{Qwen3.5-27B} over 56,939 paired query--arm evaluations. Binary agreement compares exact yes/no verdicts on \textsc{LongMemEval} and rubric pass decisions ($\text{score} \ge 60$, grounding $\ge 3$, coverage $\ge 3$) on \textsc{UtilMem}; reliability reports Cohen's $\kappa$ for binary \textsc{LongMemEval} judging and Pearson $r$ for continuous 0--100 \textsc{UtilMem} rubric scores.}
\label{tab:judge_agree}
\end{table*}

\subsection{Tri-Judge Validation, Unfiltered \textsc{UtilMem}, and Blind Cross-Family Audit}

Because \textsc{DeepSeek-V4.1-Flash} generated our synthetic training corpus and also serves as our primary unified judge, an independent reviewer could worry about generator--judge family bias. Tables~\ref{tab:judge_agree}, \ref{tab:gpt4o_anchor}, and \ref{tab:utilmem_local} directly rule out this confound by cross-checking our evaluations against two completely separate judge families: archived official \textsc{GPT-4o-2024-11-20} and deterministic local \textsc{Qwen3.5-27B}. Across 56,939 paired query--arm judgments, inter-judge agreement reaches 95.6\%--97.4\% on \textsc{LongMemEval} with chance-corrected Cohen's $\kappa = 0.893$--$0.933$, and continuous 0--100 \textsc{UtilMem} rubric scores correlate strongly at Pearson $r = 0.826$--$0.832$. Table~\ref{tab:gpt4o_anchor} confirms that every \textsc{LongMemEval} ranking replicates under the frozen \textsc{GPT-4o} judge, where \method{} beats \textsc{Rerank-TopK} by $+4.80$\,pp on 8B and $+2.80$\,pp on \textsc{GPT-4o-mini}. Across 36 manual boundary disputes, \textsc{DeepSeek} adhered more strictly to official rubrics than \textsc{GPT-4o}, winning 18.0 of 32 \textsc{LongMemEval} cases and 3.5 of 4 \textsc{UtilMem} cases.

On \textsc{UtilMem}, 10--15 of 500 queries triggered upstream HTTP 400 moderation filters in the external \textsc{DeepSeek} API and were excluded symmetrically across all systems. To confirm that those exclusions do not distort \textsc{UtilMem} conclusions, Table~\ref{tab:utilmem_local} evaluates all 500 unfiltered queries under deterministic local \textsc{Qwen3.5-27B} judging. Pooled three-reader clustered bootstrap confirms that \method{} outperforms \textsc{FacLoc} by $+7.59$ points (Holm $p < 0.0001$) and $\tau=0$ by $+1.23$ points (Holm $p = 0.0444$), while trailing \textsc{Rerank-TopK} non-significantly by $-1.58$ points (Holm $p = 0.1072$). All primary arms render chronological session blocks with identical deterministic calendar relative-gap headers specifying day offsets to the reference date and inter-session gaps (Appendix~\ref{app:prompts_cases}). Stripping these headers in the \textsc{w/o Gap Tag} control yields a modest $+1.04$-point gain on \textsc{UtilMem} (Holm $p = 0.0480$), where prompts are dominated by non-dated multi-topic synthesis rather than calendar arithmetic.

Finally, to check answer quality beyond automated scorecards, Table~\ref{tab:blind_audit} summarizes a 299-item blind cross-family audit conducted with \textsc{Claude Opus 5.5} under anonymized A/B pairing across all four benchmarks. Alongside exact accuracy and agreement with \textsc{DeepSeek}, the table reports non-tie preference rate, defined as \method{}'s win share among paired outputs where the blind auditor preferred one response over the other. Agreement with \textsc{DeepSeek} spans 95.3\%--98.7\% ($\kappa = 0.880$--$0.966$) on standard dialogue benchmarks. On \textsc{LME-S}, \method{} ties \textsc{Rerank-TopK} in binary accuracy on the 75-item subsample while achieving a 65.0\% non-tie preference rate; on \textsc{LME-M}, \method{} trends ahead by $+6.67$\,pp with a nominally significant 79.0\% preference win rate (unadjusted $p = 0.019$). On \textsc{LoCoMo}, \method{} leads \textsc{MemLoc-4B} by $+8.11$\,pp, and on \textsc{UtilMem}, \method{} exceeds \textsc{FacLoc} by $+8.77$ rubric points (unadjusted $p < 10^{-4}$). Note that 61 of 75 \textsc{UtilMem} audit references hit the 22k-character gold-session truncation cap and this audit paired \method{} against \textsc{FacLoc}, so it does not extrapolate to \textsc{Rerank-TopK}.

\begin{table*}[t]
\centering
\resizebox{\textwidth}{!}{%
\begin{tabular}{@{}llcccccccc@{}}
\toprule
\textbf{Bench.} & \textbf{Reader (Frozen \textsc{GPT-4o} Judge)} & \textbf{NaiveRAG} & \textbf{FacLoc} & \textbf{Raw ($\tau{=}.05$)} & \textbf{TopK} & \textbf{CausalMLP} & \textbf{Vernier ($\tau{=}0$)} & \textbf{Vernier} & \textbf{Oracle} \\
\midrule
\textsc{LME-S} & \textsc{Qwen3-8B}     & 66.20 & 70.60 & 54.80 & 73.20 & 77.80 & 74.40 & \textbf{78.00} & 77.40 \\
\textsc{LME-S} & \textsc{GPT-4o-mini}  & 74.40 & 77.80 & 57.40 & 82.60 & 80.80 & 81.60 & \textbf{85.40} & -- \\
\textsc{LME-M} & \textsc{Qwen3-8B}     & 60.60 & 58.20 & 50.00 & 66.00 & \textbf{70.40} & 68.20 & 69.40 & 71.20 \\
\textsc{LME-M} & \textsc{GPT-4o-mini}  & 67.20 & 69.00 & 55.00 & 74.40 & \textbf{76.80} & 73.00 & 75.60 & -- \\
\bottomrule
\end{tabular}}
\caption{Archived official \textsc{GPT-4o-2024-11-20} judge anchor accuracy (\%) over 23,797 evaluations ($n=500$ per benchmark). \textbf{Bold} marks the best non-oracle selector per row; dashes denote uncomputed \textsc{GPT-4o-mini} Oracle controls. On \textsc{LME-S} with \textsc{GPT-4o-mini}, \method{} beats \textsc{CausalMLP} by $+4.60$\,pp (Holm $p=0.030$) and \textsc{Rerank-TopK} by $+2.80$\,pp. On \textsc{LME-M}, compact \method{} at $\tau=0.08$ scores 69.80\% on 8B, and omitting date headers scores 76.60\% on \textsc{GPT-4o-mini}.}
\label{tab:gpt4o_anchor}

\vspace{7pt}
\resizebox{\textwidth}{!}{%
\begin{tabular}{@{}lccccccccc@{}}
\toprule
\textbf{Reader (Local 27B Judge, $n{=}500$)} & \textbf{NaiveRAG} & \textbf{FacLoc} & \textbf{Raw ($\tau{=}.05$)} & \textbf{TopK} & \textbf{CausalMLP} & \textbf{Vernier ($\tau{=}0$)} & \textbf{Vernier (Ours)} & \textbf{w/o Gap Tag} & \textbf{Oracle} \\
\midrule
\textsc{Qwen3-8B}    & 31.63 & 24.86 & \textbf{33.26} & 32.41 & 32.22 & 29.77 & 31.43 & 32.18 & 45.60 \\
\textsc{GPT-4o-mini} & 38.49 & 33.47 & 38.08 & 40.27 & 39.80 & 39.43 & 39.33 & \textbf{40.55} & -- \\
\textsc{Qwen3.5-27B} & 32.20 & 22.22 & 32.17 & \textbf{35.30} & 32.81 & 30.40 & 32.53 & 33.80 & -- \\
\bottomrule
\end{tabular}}
\caption{Unfiltered $n=500$ \textsc{UtilMem} 0--100 rubric scores under deterministic local \textsc{Qwen3.5-27B} judging with zero API moderation exclusions. \textbf{Bold} marks the highest non-oracle score per row; dashes denote uncomputed Oracle controls. All primary arms share symmetric calendar relative-gap headers; \textsc{w/o Gap Tag} strips those headers from \method{}. Pooled 3-reader clustered bootstrap over $n=1{,}496$ valid paired JSON verdicts confirms \method{} beats \textsc{FacLoc} ($+7.59$, Holm $p<0.0001$) and $\tau=0$ ($+1.23$, Holm $p=0.0444$) while trailing \textsc{Rerank-TopK} non-significantly ($-1.58$, Holm $p=0.1072$).}
\label{tab:utilmem_local}

\vspace{7pt}
\resizebox{\textwidth}{!}{%
\begin{tabular}{@{}lllccccccc@{}}
\toprule
\textbf{Benchmark} & \textbf{Reader} & \textbf{Paired Baseline} & \textbf{$N$} & \textbf{\method{} vs.\ Base} & \textbf{$\Delta$ ($p$)} & \textbf{Pref.\ W/T/L} & \textbf{Non-Tie Win ($p$)} & \textbf{DS Agree} & \textbf{DS $\kappa$ / $r$} \\
\midrule
\textsc{LME-S} & \textsc{Qwen3-8B} & \textsc{Rerank-TopK} & 75 & 73.3\% vs.\ 73.3\% & $0.00$\,pp ($1.00$) & 13 / 55 / 7 & 65.0\% ($0.263$) & 98.7\% & $\kappa{=}0.966$ \\
\textsc{LME-M} & \textsc{Qwen3-8B} & \textsc{Rerank-TopK} & 75 & 78.7\% vs.\ 72.0\% & $+6.67$\,pp ($0.180$) & 15 / 56 / 4 & \textbf{79.0\%} ($0.019$) & 95.3\% & $\kappa{=}0.880$ \\
\textsc{LoCoMo} & \textsc{Qwen3-8B} & \textsc{MemLoc-4B} & 74 & 68.9\% vs.\ 60.8\% & $+8.11$\,pp ($0.345$) & 21 / 42 / 11 & 65.6\% ($0.110$) & 97.3\% & $\kappa{=}0.941$ \\
\textsc{UtilMem}$^\dagger$ & \textsc{GPT-4o-mini} & \textsc{FacLoc} & 75 & 49.4 vs.\ 40.7 & \textbf{$+8.77$ pts} ($<10^{-4}$) & 48 / 11 / 16 & \textbf{75.0\%} ($<10^{-4}$) & 85.6\% & $r{=}0.716$ \\
\bottomrule
\end{tabular}}
\caption{299-item blind cross-family audit with \textsc{Claude Opus 5.5} under anonymized A/B pairing (one duplicate \textsc{LoCoMo} item deduplicated, giving $N=74$). Columns report accuracy (\%) on \textsc{LME-S/M}/\textsc{LoCoMo} or 0--100 rubric score on \textsc{UtilMem}, unadjusted two-sided exact McNemar or bootstrap $p$-value, blind win/tie/loss counts, non-tie preference rate with unadjusted two-sided sign-test $p$, and agreement with primary \textsc{DeepSeek} (DS) judging. \textbf{Bold} marks nominally significant unadjusted preference or score margins ($p<0.05$). $^\dagger$On \textsc{UtilMem}, 61/75 gold references hit the 22k-character truncation cap and pairing was against \textsc{FacLoc}, so this result does not extrapolate to \textsc{Rerank-TopK}.}
\label{tab:blind_audit}
\end{table*}
\FloatBarrier

\clearpage
\section{Verbatim Prompt Templates, Error Decomposition, and Qualitative Case Analysis}
\label{app:prompts_cases}

\subsection{Verbatim Prompt Templates}

To ensure exact reproducibility, Boxes~E1--E4 reproduce the verbatim selector input format, symmetric session-header scheme, reader prompts, automated judge prompts, and synthetic dialogue generation template. Every compared query-time selector receives identical reader prompts and identical session headers so QA differences reflect only evidence selection.

\begin{promptbox}{Box E1: \method{} Single-Prefill Selector Input and Symmetric Calendar Session Headers}
\textbf{Selector System Prompt:}\\
\emph{You evaluate candidate conversational memory turns jointly as an evidence set. Identify which turns are needed to answer the question and which pairs are complementary (state updates, multi-hop bridges, multi-session counts) versus redundant distractors.}\\[3pt]
\textbf{Unindexed User Input Format (zero numeric rank IDs; each candidate capped at 72 tokens):}\\
\texttt{[Reference Date: \{q\_date\}]}\\
\texttt{Question: \{question\}}\\
\texttt{Candidates:}\\
\texttt{[\{date\_hdr\} | \{speaker\}] \{turn\_text\} </C>}\\
\texttt{...} \emph{(repeated for all $M \le 32$ turns)} \texttt{<S>}\\[3pt]
\textbf{Symmetric Reader Session Header (applied identically to all primary arms):}\\
\texttt{=== Conversation Session [\{date\_hdr\} | Offset: \{d\} days (\{w\} weeks) before Reference Date | +\{g\} days after previous session] ===}\\
\emph{Implementation note: in \textsc{Qwen3}, \texttt{</C>} uses single token \texttt{<|vision\_pad|>} (fallback \texttt{>{}>}), \texttt{<S>} reads the final token of \texttt{<|im\_end|>\textbackslash n<|im\_start|>assistant\textbackslash n}, and inter-session gaps $\ge 7$ days also append \texttt{(\{w\} weeks)}. Control \textsc{w/o Date} / \textsc{w/o Gap Tag} strips both offset/gap clauses and inline resolved calendar date tags.}
\end{promptbox}

\begin{promptbox}{Box E2: Verbatim Symmetric Reader Prompts Across Benchmarks}
\textbf{1. \textsc{LongMemEval-S/M} (User message):}\\
\emph{Answer the question using the provided conversation evidence. When combining, comparing, or calculating information is necessary, you may include brief relevant evidence and calculations. Check that the facts you use satisfy the conditions in the question and that your conclusion agrees with those facts. If the evidence is insufficient, say what is missing rather than guessing.}\\
\texttt{Conversation evidence: \{evidence\}}\\
\texttt{Question reference date: \{question\_date\}}\\
\texttt{Question: \{question\} \textbackslash n Answer:}\\[4pt]
\textbf{2. \textsc{LoCoMo-1986} (System + User messages):}\\
\textbf{System:} \emph{You are a precise conversational memory Q\&A assistant. Answer concisely based strictly on the provided conversation sessions. Output only the final answer, with no reasoning, JSON, or wrapper text.}\\
\textbf{User:} \emph{Answer the question concisely based strictly on the conversation sessions below. Guidelines: (1) State the direct factual answer concisely (names, items, places, or short phrases; separate multiple items with commas). (2) When asked when something happened or given relative time expressions (such as `yesterday', `last week', `next month', or `last year'), always output the resolved concrete calendar date, period, month, or year (using the session dates and inline date tags, e.g., `7 May 2023', `June 2023', `2022')---never answer with bare relative words. (3) If the provided conversation sessions do not contain enough information to answer the question (or ask about something never mentioned), reply strictly with: `Not mentioned in the conversation.' (4) Return ONLY the concise final answer with no explanation.}\\
\texttt{Context: \{evidence\} \textbackslash n Question: \{question\} \textbackslash n Answer:}\\[4pt]
\textbf{3. \textsc{UtilMem} (System + User messages):}\\
\textbf{System:} \emph{You are an expert personal assistant answering complex user requests by synthesizing distributed facts, constraints, preferences, and updates across multi-session conversation history.}\\
\textbf{User:} \emph{Answer the user's request thoroughly and accurately using ONLY the relevant facts, timelines, preferences, and constraints supported by the conversation evidence below. Guidelines: (1) Synthesize all relevant details across multiple sessions coherently and reconcile any updates chronologically. (2) Stay strictly grounded in the evidence---ignore unrelated distractor topics and do not hallucinate unsupported facts. (3) Provide a clear, well-structured, directly actionable response.}\\
\texttt{Conversation Evidence: \{evidence\} \textbackslash n User Request: \{question\} \textbackslash n Response:}
\end{promptbox}

\begin{promptbox}{Box E3: Verbatim Automated Judge Prompts (Shared Across All Three Judge Families)}
\textbf{1. \textsc{LongMemEval-S/M} (Official upstream \texttt{get\_anscheck\_prompt}):}\\
\emph{I will give you a question, a correct answer, and a response from a model. Please answer yes if the response contains the correct answer. Otherwise, answer no. If the response is equivalent to the correct answer or contains all the intermediate steps to get the correct answer, you should also answer yes. If the response only contains a subset of the information required by the answer, answer no. [Temporal clause: do not penalize off-by-one errors for the number of days/weeks/months. Update clause: previous info alongside the updated answer is correct if the updated answer matches. Preference clause: satisfies the desired personalized response rubric. Abstention clause: yes if the model identifies the question as unanswerable.] Question: \{q\} / Correct Answer: \{gold\} / Model Response: \{resp\} / Answer yes or no only.}\\[4pt]
\textbf{2. \textsc{LoCoMo-1986} Cat 1--4 (Official \texttt{ACCURACY\_PROMPT}) and Cat-5 Abstention:}\\
\textbf{Cat 1--4:} \emph{Your task is to label an answer to a question as `CORRECT' or `WRONG' given (1) a question, (2) a gold answer, and (3) a generated answer. Be generous with phrasing and date format equivalence as long as the response refers to the same topic, date, or time period as the gold answer. Return JSON with key \texttt{"label"} (`CORRECT' or `WRONG').}\\
\textbf{Cat-5:} \emph{You are evaluating an AI assistant on an UNANSWERABLE / ADVERSARIAL conversational memory question. Given Question, Known False Trap Answer, and Model Response: label CORRECT if the model clearly abstains or states the information is missing; label WRONG if it hallucinates or echoes the trap answer. Return JSON with keys \texttt{"reason"} and \texttt{"label"}.}\\[4pt]
\textbf{3. \textsc{UtilMem} (Oracle-Gold-Session-Grounded 0--100 Rubric; reference capped at 22k chars):}\\
\emph{You are an expert, strictly impartial evaluator grading an AI assistant's response given User Request, Ground-Truth Reference Sessions, and Assistant Response. Score 0--5 integers for: (1) \texttt{grounding} (strict support, zero hallucination), (2) \texttt{coverage} (incorporates distributed facts, constraints, updates), (3) \texttt{synthesis} (coherent, actionable integration), (4) \texttt{distractor\_resistance} (ignores irrelevant distractors). Compute integer \texttt{overall\_score} (0--100) and binary \texttt{pass} (1 iff $\text{overall\_score}\ge 60$, $\text{grounding}\ge 3$, $\text{coverage}\ge 3$). Return strict JSON.}
\end{promptbox}

\begin{promptbox}{Box E4: Synthetic Longitudinal Dialogue Generation Prompt (\textsc{DeepSeek-V4.1-Flash})}
\textbf{System:} \emph{You are an expert longitudinal conversational memory dataset architect. Return strictly valid JSON matching the requested schema. Never leak internal IDs (such as S0, Ch1, t3) inside dialogue utterances, questions, or answers.}\\[3pt]
\textbf{User Blueprint (Pass 1, 4 chapters $\times$ 8 sessions $\times$ 7--9 turns per long history):}\\
\emph{Generate Chapter \{c\}/4 of longitudinal conversation \{conv\_id\} (domain: \{domain\}; speakers: \{speakers\}; dated sessions S0--S7; four evolving arcs A--D with Chapter 3 introducing an explicit state update/pivot), plus 2 intra-chapter QAs. Mix ~70\% short 12--42-word turns (with intra-turn coreference, 2-turn elliptical Q\&A, and relative dates) and ~30\% structured 110--210-word turns (1--2 buried facts amid detachable tips/boilerplate) so length heuristics fail. Include realistic phatic and rejected-plan distractors. Return JSON keys \texttt{chapter\_summary}, \texttt{sessions}, and \texttt{intra\_qas}. Pass 2 consumes all 32 session summaries to generate 8 cross-chapter QAs covering all eight regimes including 10\% false-premise abstention.}
\end{promptbox}

\subsection{Stage-by-Stage Error and 72-Token Truncation Decomposition}

\begin{figure*}[t]
    \centering
    \includegraphics[width=0.98\textwidth]{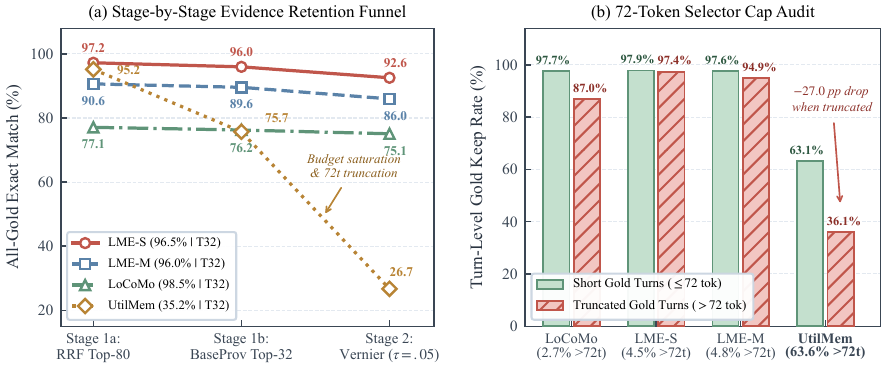}
    \caption{Stage-by-stage evidence retention and 72-token scoring truncation audit across all four benchmarks (complete breakdown in Table~\ref{tab:error_decomp}). Left: All-Gold EM (\%) from Stage-1a RRF Top-80 through Stage-1b \textsc{BaseProvence} Top-32 to deployed \method{} ($\tau=0.05$); legend entries report conditional retention given Top-32 hits (\method{} $|$ T32), remaining at 96.0\%--98.5\% on standard dialogue benchmarks before dropping to 35.2\% on dense OOD \textsc{UtilMem}, where upstream annotations specify whole evidence sessions so EM measures all-gold-session coverage. Right: turn-level gold keep rate for short turns ($\le 72$ tokens) versus truncated turns ($>72$ tokens), showing a $-27.0$\,pp drop under \textsc{UtilMem}'s session-derived positive turn proxy where 63.6\% of session-matched turns exceed the selector's 72-token scoring cap.}
    \label{fig:app_funnel}
\end{figure*}

When an end-to-end answer is wrong, the error can arise from three distinct stages: Stage-1a/1b front-end retrieval missing a gold turn before \method{} sees it, Stage~2 pruning a recoverable gold turn inside Top-32, or the downstream reader failing despite receiving complete evidence. To localize where losses happen and quantify the cost of our 72-token selector scoring cap, Figure~\ref{fig:app_funnel} and Table~\ref{tab:error_decomp} trace All-Gold EM across every stage alongside turn-level keep rates for short versus truncated turns and conditional \textsc{Qwen3.5-27B} QA accuracy.

Two sharply different regimes emerge. On standard conversational benchmarks, \method{}'s conditional retention given Top-32 hits (\method{} $|$ T32) stays between 96.0\% and 98.5\%, and gold turns exceeding 72 tokens are rare at 2.7\%--4.8\%. Remaining misses stem primarily from the frozen Top-32 front-end ceiling, particularly on \textsc{LoCoMo} where 365 of 1,536 queries miss a gold turn before Stage~2, whereas \method{} drops a recoverable Top-32 gold turn on only 16--18 queries per benchmark. When all gold turns survive, 27B accuracy stays high at 85.3\%--89.0\%, confirming that reader performance tracks complete evidence retention directly.

Dense OOD \textsc{UtilMem} exposes the opposite bottleneck. Because upstream \textsc{UtilMem} specifies designated evidence sessions averaging 4.76 sessions per query without turn-level necessity labels \citep{qing2026utilmem}, stage All-Gold EM tracks full-session coverage across all designated sessions rather than turn minimality, while the 72-token audit treats all 23,285 Top-32 turns belonging to those sessions as a session-derived positive turn proxy. Under this proxy, 99.4\% of queries contain at least one session-matched turn exceeding 72 tokens, and 63.6\% of those turns are truncated during selector scoring. Combined with 94.1\% budget binding at $B=1000$, proxy turn keep rate drops by $-27.0$\,pp from 63.1\% on short turns to 36.1\% on truncated turns, pulling conditional all-session retention down to 35.2\%. Measuring mean session recall directly on the same packed predictions clarifies partial coverage: across all 1,717 queries, mean session recall moves from 99.06\% at Top-80 and 94.19\% at Top-32 to \textbf{73.24\%} for \method{} versus 61.18\% for \textsc{Rerank-TopK} (+12.06\,pp; 72.89\% versus 62.28\% on the 500-query QA subset, alongside 26.80\% versus 14.40\% all-session EM). Whenever all designated sessions are covered, 27B rubric score reaches 41.7 versus 32.7 when \method{} drops a session and 28.5 when Top-32 misses one, pinpointing why dense-synthesis QA gains remain narrow even when partial session coverage improves.

\begin{table*}[t]
\centering
\resizebox{\textwidth}{!}{%
\begin{tabular}{@{}l|cccc|cc|ccc@{}}
\toprule
\multirow{2}{*}{\textbf{Benchmark ($n_{\text{ans}}$)}} & \multicolumn{4}{c|}{\textbf{Stage All-Gold EM (\%)}} & \multicolumn{2}{c|}{\textbf{72-Token Cap Audit}} & \multicolumn{3}{c}{\textbf{27B QA Conditional on Evidence}} \\
 & \textbf{Top-80} & \textbf{Top-32} & \textbf{\method{}} & \textbf{\method{} $|$ T32} & \textbf{Gold $>72$t} & \textbf{Keep $\le$72 / $>$72t} & \textbf{All Kept ($n$)} & \textbf{Drop ($n$)} & \textbf{T32 Miss ($n$)} \\
\midrule
\textsc{LoCoMo} (1,536) & 77.08 & 76.24 & 75.07 & 98.46 & 2.66\% & 97.7\% / 87.0\% & 85.3\% (1,153) & 61.1\% (18) & 51.5\% (365) \\
\textsc{LME-S} (470)    & 97.23 & 95.96 & 92.55 & 96.45 & 4.53\% & 97.9\% / 97.4\% & 89.0\% (435)   & 56.3\% (16) & 36.8\% (19)  \\
\textsc{LME-M} (470)    & 90.64 & 89.57 & 85.96 & 95.96 & 4.77\% & 97.6\% / 94.9\% & 87.1\% (404)   & 41.2\% (17) & 18.4\% (49)  \\
\textsc{UtilMem}$^\dagger$ (1,717)& 95.17 & 75.71 & 26.67 & 35.23 & 63.59\% & 63.1\% / 36.1\% & 41.7 (133)    & 32.7 (249)  & 28.5 (109)   \\
\bottomrule
\end{tabular}}
\caption{Stage-by-stage error and 72-token candidate truncation decomposition across all four benchmarks ($n_{\text{ans}}$ = answerable queries with annotated gold evidence; 4 of 1,540 \textsc{LoCoMo} Cat 1--4 queries lack upstream gold turn IDs and are evaluated only in end-to-end QA). Stage columns track All-Gold EM (\%) after Stage-1a RRF Top-80, Stage-1b \textsc{BaseProvence} Top-32, and \method{} ($\tau=0.05$, $k_{\min}=2$), plus conditional retention given Top-32 hits (\method{} $|$ T32). Truncation columns report the percentage of gold turns exceeding 72 tokens and turn-level keep rates for short versus truncated turns. Right columns report \textsc{Qwen3.5-27B} \textsc{DeepSeek} accuracy (\%) or \textsc{UtilMem} 0--100 rubric score (matched $n=491$) partitioned by whether all gold evidence is kept, dropped by \method{}, or missed by Top-32. $^\dagger$Upstream \textsc{UtilMem} annotates designated evidence sessions without turn-level necessity labels, so Stage EM measures all-gold-session coverage (mean session recall: 99.06\% Top-80, 94.19\% Top-32, 73.24\% \method{} vs.\ 61.18\% \textsc{Rerank-TopK}; 72.89\% vs.\ 62.28\% on the 500-query QA subset), while the 72-token audit treats all 23,285 Top-32 turns inside gold sessions as a session-derived positive turn proxy.}
\label{tab:error_decomp}
\end{table*}

\subsection{Qualitative Case Analysis: Wins and Failure Modes}

To illustrate concretely how answer-oriented evidence selection differs from relevance-based retrieval in practice, Table~\ref{tab:cases} contrasts three representative \method{} wins covering temporal reasoning, multi-hop entity bridging, and knowledge updates under noise against two genuine failure modes.

\paragraph{Case 1 (Win, Temporal): \textsc{LongMemEval-S} \texttt{4dfccbf7}.}
As illustrated in Figure~\ref{fig:motivated}, computing the 24-day span between Feb~1 ukulele lessons ($g_1$) and a Feb~25 guitar repair ($g_2$) requires separating an earlier Feb~1 plan ($t_1$, $\cos=0.726$). Full-context reading over 111,180 tokens predicts \emph{``0 days''}, isolated pointwise thresholding collapses both gold scores below $0.001$ and triggers abstention, and \textsc{FacLoc} repels $g_2$ due to lexical overlap with $t_1$. \textsc{Rerank-TopK} retains both golds inside 31 turns totaling 1,885 tokens, but \textsc{Qwen3-8B} still outputs \emph{``0 days''}. \method{} scores both golds at $0.99$, prunes 27 noise turns, and packs 5 turns in 404 tokens so both 8B and 27B readers answer \emph{``24 days''}.

\paragraph{Case 2 (Win, Multi-Hop): \textsc{LoCoMo} \texttt{conv-41:018}.}
Answering \emph{``Who did John go to yoga with?''} across a 25.7k-token dialogue requires linking two topically overlapping turns ($\cos=0.790$). Full-context reading confuses entity names and answers \emph{``Maria''}, similarity-repulsive \textsc{FacLoc} drops the turn naming \emph{Rob}, and iterative \textsc{MemLoc-4B} returns generic \emph{``a colleague''} at 2,910 tokens. \method{} retains both complementary turns at scores $0.97$ and $0.78$ within 659 tokens, enabling both readers to answer \emph{``Rob''}.

\paragraph{Case 3 (Win, Update): \textsc{LongMemEval-S} \texttt{852ce960}.}
On December~18, 2023, the user asks \emph{``What was the amount I was pre-approved for when I got my mortgage from Wells Fargo?''}, with updated gold answer \emph{\$400,000}. An earlier August~11 session records an initial \emph{\$350,000} pre-approval before a later session updates the approved figure to \emph{\$400,000}; \textsc{LongMemEval} marks both timeline turns as gold. \textsc{Rerank-TopK} retrieves both turns alongside 13 unrelated housing and finance distractors, filling 18 turns in 1,258 tokens. Buried inside that noisy packet, \textsc{Qwen3-8B} latches onto the earlier \emph{\$350,000} figure and answers incorrectly. Comparing all 32 candidates jointly, \method{} suppresses all 13 noise turns below $\tau=0.05$ and packs a compact 5-turn chronological timeline totaling 586 tokens, a 53\% reduction, preserving both update turns ($p=0.971$ and $p=0.114$). With distracting turns cleared away, \textsc{Qwen3-8B} tracks the chronological update cleanly and outputs \emph{``\$400,000''}, matching 27B. The same noise-pruning flip recurs on \textsc{LME-S} \texttt{b6019101}, where trimming a 21-turn, 1,364-token \textsc{Rerank-TopK} packet down to 8 turns in 629 tokens flips \textsc{Qwen3-8B} from an outdated count of \emph{4} to the updated \emph{5} Marvel films.

\paragraph{Case 4 (Fail, Cutoff): \textsc{LoCoMo} \texttt{conv-41:042}.}
Answering \emph{``What area was hit by a flood?''}, with gold answer \emph{West County}, requires bridging two turns: a 32-token turn \emph{``My old area was hit by a nasty flood last week''} scored at $p=0.975$ and a 23-token turn sharing zero lexical overlap with \emph{flood}, \emph{``I want to work on improving my old area, West County, too''}. \method{} assigns the second turn $p=0.0415$, just $0.0085$ below fixed $\tau=0.05$, and drops it. Packing 532 tokens without the entity bridge causes both 8B and 27B readers to answer \emph{``John's old area''}, whereas full-budget \textsc{Rerank-TopK} retains both turns in 1,448 tokens and answers \emph{``West County''} on both readers. This failure motivates query-conditioned thresholds in future work.

\paragraph{Case 5 (Fail, Truncation): \textsc{UtilMem} \texttt{fitness::q0}.}
A dense fitness-planning request (\texttt{multi\_0011\_04b92578::fitness::q0}) asks the assistant to reconcile workout progression, creatine intake, protein targets, and back/shoulder constraints across seven gold sessions. Five gold sessions enter Top-32, but 17 of the 18 turns from those sessions exceed the 72-token selector cap. At $B=1000$, both \method{} and \textsc{Rerank-TopK} saturate the budget with seven turns near 1,375 prompt tokens. Because truncated scoring undervalues long multi-topic turns, \method{} covers only 3 of 7 gold sessions, scoring 12 on 8B and 30 on 27B, trailing \textsc{Rerank-TopK}, which covers 4 of 7 sessions and scores 50 on 8B and 78 on 27B.

\begin{table*}[t]
\centering
\resizebox{\textwidth}{!}{%
\begin{tabular}{@{}llcccc@{}}
\toprule
\textbf{Case (Regime / Gold)} & \textbf{System} & \textbf{Reader Tok} & \textbf{Gold Kept} & \textbf{\textsc{Qwen3-8B} Output} & \textbf{\textsc{Qwen3.5-27B} Output} \\
\midrule
\multirow{5}{*}{\shortstack[l]{\textbf{Case 1 (Win):} \textsc{LME-S} \texttt{4dfccbf7}\\111.2k hist., $\cos{=}0.726$\\Gold: \textbf{24 days}}}
 & Native 262k \textsc{FullContext} & 111,180 & -- & -- & \emph{``0 days''} ($\times$) \\
 & \textsc{Raw Reranker} ($\tau{=}.05$) & 251 & 0 / 2 & \emph{``not possible''} ($\times$) & \emph{``impossible''} ($\times$) \\
 & \textsc{Similarity Rep.\ (FacLoc)} & 1,874 & 1 / 2 (drops $g_2$) & \emph{``0 days''} ($\times$) & \emph{``0 days''} ($\times$) \\
 & \textsc{Rerank-TopK} & 1,885 & 2 / 2 (31 turns) & \emph{``0 days''} ($\times$) & \emph{``24 days''} (\checkmark) \\
 & \textbf{\method{} ($\tau{=}0.05$)} & \textbf{404} & \textbf{2 / 2 (5 turns)} & \textbf{\emph{``24 days''}} (\checkmark) & \textbf{\emph{``24 days''}} (\checkmark) \\
\midrule
\multirow{4}{*}{\shortstack[l]{\textbf{Case 2 (Win):} \textsc{LoCoMo} \texttt{conv-41:018}\\25.7k hist., $\cos{=}0.790$\\Gold: \textbf{Rob}}}
 & \textsc{FullContext} & 25,697 & -- & \emph{``Maria''} ($\times$) & -- \\
 & \textsc{Similarity Rep.\ (FacLoc)} & 1,278 & 1 / 2 (drops \emph{Rob}) & \emph{``Max''} ($\times$) & \emph{``Not mentioned''} ($\times$) \\
 & \textsc{MemLoc-4B} (Official) & 2,910 & Misses \emph{Rob} & \emph{``a colleague''} ($\times$) & \emph{``a colleague''} ($\times$) \\
 & \textbf{\method{} ($\tau{=}0.05$)} & \textbf{659} & \textbf{2 / 2 ($p{=}.97,.78$)} & \textbf{\emph{``Rob''}} (\checkmark) & \textbf{\emph{``Rob''}} (\checkmark) \\
\midrule
\multirow{2}{*}{\shortstack[l]{\textbf{Case 3 (Win):} \textsc{LME-S} \texttt{852ce960}\\Update (\$350k $\rightarrow$ \$400k), Gold: \textbf{\$400,000}}}
 & \textsc{Rerank-TopK} & 1,258 & 2 / 2 (18 turns) & \emph{``\$350,000''} ($\times$) & \emph{``\$400,000''} (\checkmark) \\
 & \textbf{\method{} ($\tau{=}0.05$)} & \textbf{586} & \textbf{2 / 2 (5 turns)} & \textbf{\emph{``\$400,000''}} (\checkmark) & \textbf{\emph{``\$400,000''}} (\checkmark) \\
\midrule
\multirow{2}{*}{\shortstack[l]{\textbf{Case 4 (Fail):} \textsc{LoCoMo} \texttt{conv-41:042}\\Cutoff ($p{=}.0415$), Gold: \textbf{West County}}}
 & \textsc{Rerank-TopK} & 1,448 & \textbf{2 / 2} & \textbf{\emph{``West County''}} (\checkmark) & \textbf{\emph{``West County''}} (\checkmark) \\
 & \method{} ($\tau{=}0.05$) & 532 & 1 / 2 (drops bridge) & \emph{``John's old area''} ($\times$) & \emph{``John's old area''} ($\times$) \\
\midrule
\multirow{2}{*}{\shortstack[l]{\textbf{Case 5 (Fail):} \textsc{UtilMem} \texttt{fitness::q0}\\7 gold sess., 17/18 $>72$t (0--100 rubric)}}
 & \textsc{Rerank-TopK} & 1,359 / 1,375 & \textbf{4 / 7 sess.} (7 turns) & \textbf{50 / 100} & \textbf{78 / 100} \\
 & \method{} ($\tau{=}0.05$) & 1,366 / 1,384 & 3 / 7 sess.\ (7 turns) & 12 / 100 & 30 / 100 \\
\bottomrule
\end{tabular}}
\caption{Side-by-side qualitative walkthroughs covering three representative \method{} wins (Cases 1--3) and two genuine failure modes (Cases 4--5). \textbf{Bold} marks the stronger system per case; \checkmark{} and $\times$ indicate correct versus incorrect \textsc{DeepSeek} verdicts, dashes denote unrun reader cells, and Case 5 reports 8B / 27B prompt tokens and 0--100 rubric scores.}
\label{tab:cases}
\end{table*}
\FloatBarrier

\end{document}